\documentclass[default,iicol]{sn-jnl}

\usepackage{graphicx}%
\usepackage{multirow}%
\usepackage{amsmath,amssymb,amsfonts}%
\usepackage{amsthm}%
\usepackage{mathrsfs}%
\usepackage[title]{appendix}%
\usepackage{xcolor}%
\usepackage{textcomp}%
\usepackage{manyfoot}%
\usepackage{booktabs}%
\usepackage{algorithm}%
\usepackage{algorithmicx}%
\usepackage{algpseudocode}%
\usepackage{listings}%

\usepackage[sorting=none, maxbibnames=99]{biblatex}
\usepackage{nomencl}

\usepackage{subcaption}
\usepackage{tikz}

\usepackage{array}
\usepackage{colortbl}
\usepackage{tabulary}
\usepackage[normalem]{ulem}

\makeatletter
\newif\if@checkentries
\renewenvironment{nomenclature}[1][]
 {\if\relax\detokenize{#1}\relax
    \@checkentriesfalse
  \else
    \settowidth{\@tempdima}{#1\quad}%
    \@checkentriestrue
  \fi
  \def\entry##1##2{%
    \sbox\@tempboxa{##1\quad}%
    \if@checkentries
      \ifdim\wd\@tempboxa>\@tempdima
        \@latex@warning{Entry `\unexpanded{##1}' is wider}%
        \@tempdima=\wd\@tempboxa
      \fi
    \else
      \@tempdima=\wd\@tempboxa
    \fi
    \hangindent\@tempdima
    \noindent\makebox[\@tempdima][l]{##1}\ignorespaces##2\par}%
    \section*{Nomenclature}}
 {\par\addvspace{12pt}}
\makeatother

\theoremstyle{thmstyleone}%
\theoremstyle{thmstyletwo}%

\theoremstyle{thmstylethree}%

\makeatletter
\renewcommand\normalsize{%
  \@setfontsize\normalsize{9pt}{11pt}
  \abovedisplayskip 8pt plus 2pt minus 2pt
  \abovedisplayshortskip 6pt plus 2pt
  \belowdisplayshortskip 6pt plus 2pt minus 2pt
  \belowdisplayskip \abovedisplayskip
}
\makeatother

\DeclareMathSizes{9}{9}{7}{6}  

\begin{document}

\title[Article Title]{Digital Twin-based Control Co-Design of Full Vehicle Active Suspensions via Deep Reinforcement Learning}


\author[1]{\fnm{Ying-Kuan} \sur{Tsai}}\email{yingkuan.tsai@northwestern.edu}

\author[1]{\fnm{Yi-Ping} \sur{Chen}}\email{chenyp@u.northwestern.edu}

\author[1]{\fnm{Vispi} \sur{Karkaria}}\email{vispikarkaria2026@u.northwestern.edu}

\author*[1]{\fnm{Wei} \sur{Chen}}\email{weichen@northwestern.edu}

\affil[1]{\orgdiv{Department of Mechanical Engineering}, \orgname{Northwestern University}, \orgaddress{\city{Evanston}, \postcode{60208}, \state{IL}, \country{USA}}}




\abstract{Active suspension systems are critical for enhancing vehicle comfort, safety, and stability, yet their performance is often limited by fixed hardware designs and control strategies that cannot adapt to uncertain and dynamic operating conditions. Recent advances in digital twins (DTs) and deep reinforcement learning (DRL) offer new opportunities for real-time, data-driven optimization across a vehicle’s lifecycle. However, integrating these technologies into a unified framework for co-optimizing physical and control systems remains an open challenge. This work presents a DT-based control co-design (CCD) framework for full-vehicle active suspensions using multi-generation design concepts. Through integrating automatic differentiation into DRL, we jointly optimize physical components of suspension systems and control policies under varying driver behaviors and environmental uncertainties. The DRL technique also addresses the challenge of partial observability, where only limited states can be sensed and fed back to the controller, by learning optimal control actions directly from available sensor information. The framework incorporates model updating with quantile learning to quantify data uncertainty, enabling real-time decision-making and adaptive learning from digital-physical interactions. The approach demonstrates personalized optimization of autonomous suspension systems under two distinct driving settings (mild and aggressive). The results show that the optimized systems achieve smoother trajectories and reduce control efforts by approximately 43\% and 52\% for mild and aggressive, respectively, while maintaining ride comfort and stability. Contributions of this work include: (1) developing a DT-enabled CCD framework integrating DRL and uncertainty-aware model updating for full-vehicle active suspensions, (2) introducing a multi-generation design framework for self-improving systems across the whole lifecycle, and (3) demonstrating personalized optimization of active suspension systems for distinct types of drivers.}




\maketitle

\begin{nomenclature}[XXX] 
\entry{AI}{Artificial Intelligence}
\entry{CCD}{Control Co-Design}
\entry{DNN}{Deep Neural Network}
\entry{DRL}{Deep Reinforcement Learning}
\entry{DT}{Digital Twin}
\entry{PPO}{Proximal Policy Optimization}
\entry{RL}{Reinforcement Learning}
\entry{UQ}{Uncertainty Quantification}
\end{nomenclature}

\section{Introduction}\label{sec1}

\subsection{Motivation}

Suspension systems are central to vehicle {safety}, {ride comfort}, and {handling stability}~\cite{sun2020advanced,goodarzi2017vehicle}. They directly shape how road excitations are transmitted to the cabin, influencing passenger comfort and fatigue, while also governing tire--road contact that affects braking distance, cornering capability, and overall vehicle controllability. However, traditional passive and semi-active suspensions are typically tuned for nominal operating conditions and therefore struggle under highly uncertain and variable scenarios, including diverse road roughness, tire conditions, payload changes, and heterogeneous driver behaviors~\cite{soliman2021semi,lee2022high,qiu2025integrating}. As operating environments deviate from the design point, these systems exhibit performance degradation, either by transmitting excessive vibrations to the occupants or by allowing excessive body motion that compromises safety and control~\cite{tsai2024surrogate,weaver2020parametric}.

To overcome these structural limitations, active suspension systems, which use actuators to actively inject energy and exert an adaptive counter-force to suppress vibrations, have been developed to provide high flexibility and improved control over vehicle dynamics. This technological shift has coincided with the emergence of learning-based control methods, particularly Deep Reinforcement Learning (DRL), which are ideally suited for addressing the complexity and nonlinearity of suspension control problems~\cite{fares2020online,dridi2023new,ming2020semi,dridi2025optimizing,lee2022deep,wang2024deep}. While such approaches improve closed-loop adaptability, they remain fundamentally constrained by the {fixed} physical design of the suspension hardware (springs, dampers, and actuator characteristics). This motivates {Control Co-Design} (CCD), a design paradigm that simultaneously optimizes the physical system and the control strategy to achieve system-level optimality~\cite{garcia2019control,allison2014special,fathy2001coupling,cui2021reliability,sato2025computational}. Existing CCD studies have demonstrated significant performance improvements compared with traditional sequential (design-then-control) approaches~\cite{Bayat2025Control,allison2014co,tsai2026parametric}. However, they often treat environmental conditions and user behaviors as exogenous or static, and thus fail to embed these factors within the co-design loop. The result is limited adaptability when the operating context shifts.

{Digital twins (DTs)} offer a path beyond these limitations by enabling real-time data collection, model updating, and predictive decision-making through a continuously synchronized virtual replica of the physical system~\cite{semeraro2021digital,national2023foundational,thelen2022comprehensive,thelen2023comprehensive,karkaria2024towards,karkaria2025ai,karkaria2025optimization,chen2025real}. Crucially, DTs also facilitate \emph{personalized} optimization by learning the unique characteristics of individual assets and users~\cite{hu2023special,thelen2022comprehensive}. In the context of vehicle suspensions, this capability is particularly critical, as uncertainties stem from both environmental variability (e.g., road profiles, friction conditions, and weather) and user-specific driving behaviors (e.g., acceleration, speed, and steering inputs). Moreover, these sources of uncertainty are inherently dynamic and evolve unpredictably over time.

To effectively handle such evolving and unstructured information, the integration of Artificial Intelligence (AI)-driven learning into DTs is indispensable. AI enables DTs to process heterogeneous sensor data, infer latent system states, and continuously update predictive models, transforming the DT from a static simulation tool into an adaptive, decision-making agent. For autonomous systems such as active suspensions, this integration allows the DT to identify patterns in vehicle-road interactions and to real-time adjust the operation commands and actions. Consequently, the fusion of AI and DT technologies bridges data and design, facilitating co-optimization that enhances the adaptability, robustness, and personalization of next-generation vehicular systems~\cite{asmat2025digital}.


Despite this potential, most DT implementations in the suspension domain focus on {monitoring} and {control} alone~\cite{rosa2024digital,qiu2025integrating,li2025digital}, while {physical design integration} within the DT feedback loop remains rare~\cite{van2023digital,tsai2025multi,tsai2025digital}. Without coupling the hardware design with the continuously updated virtual model and data-driven control, adaptability is fundamentally limited: controllers must compensate for hardware that may be poorly matched to changing contexts, and design decisions cannot leverage the rich information provided by in-service data. This gap motivates a DT-enabled CCD framework that unifies design, control, and learning, thereby enabling suspensions that are not only intelligent and adaptive, but also \emph{co-evolve} physically and algorithmically over the vehicle's lifecycle.

\subsection{Multi-Generation Digital Twin Concept}

The concept of {multi-generation design} has traditionally been applied in product lifecycle engineering, remanufacturing, and sustainable design, where feedback from previous product generations informs the development of improved successors~\cite{go2015multiple,karkaria2025digital,van2023digital,asif2021methodological,nag2022multiple}. In this work, the concept is extended to the co-design of {vehicle suspension systems and controllers}, forming an evolving design-control ecosystem in which knowledge from each deployment cycle is systematically assimilated into the DT and used to guide subsequent redesigns. 
Through this mechanism, the DT serves as a persistent learning agent that accumulates operational knowledge over time, continually improving its predictive accuracy and enabling progressively better physical and control designs across generations.

Our prior work~\cite{tsai2025digital} demonstrated that the integration of DTs with CCD and DRL can significantly enhance the adaptability of active suspension systems. In particular, a quarter-car suspension case study showed that the proposed multi-generation digital twin framework enables the system to become more robust to varying initial conditions and uncertain road roughness. By continuously updating the digital model and co-optimizing the physical parameters and control policy, each successive generation achieved superior performance under broader uncertainty conditions. However, the previous study only considered a simplified suspension configuration that did not capture the complex interactions and couplings present in a full-vehicle system, and motivates us to work on more complex, realistic, and scaled testbeds.

Extending the framework to a \emph{full-vehicle} active suspension system introduces several new challenges that necessitate further methodological developments. First, the system is {partially observable}, meaning that not all states are directly measurable from onboard sensors. This requires the policy to infer unobserved dynamics through learned correlations. Second, the problem involves a {high-dimensional state and action space}, as the full vehicle includes coupled lift, pitch, and roll motions with four actuators that must be coordinated in real time. Third, the optimization must ensure {system-level dynamic controllability} across varying driving and environmental conditions, balancing comfort, stability, and energy efficiency under multiple operating modes.

\subsection{Research Objectives and Contributions}

By addressing the challenges of full-vehicle modeling, partial observability, coupled dynamics with proactive control, the DT framework extends beyond traditional offline control design toward a data-driven, learning-based, and self-improving paradigm. Through iterative model updating, uncertainty quantification, and RL-based co-optimization, the framework continuously redesigns both the physical system and its controller for future generations, enabling adaptive, personalized, and sustainable vehicle performance throughout the system lifecycle. The objective of this work is to present a multi-generation DT framework to co-optimize the physical components and active controllers for full-vehicle active suspension systems by integrating automatic differentiation with CCD and DRL techniques. This framework also enables personalized optimization by leveraging the capabilities of real-time monitoring, updating, and decision-making from DTs and allows suspension systems to continuously learned from data and to be customized for different drivers and operating conditions.

This research makes the following key contributions:
\begin{itemize}
    \item We propose a DT-enabled CCD methodology that jointly optimizes full-car suspension components and learning-based control policies within a unified DRL framework.
    \item We use a multi-generation design framework integrated with CCD formulations and the DT technology to enable continuous adaptation and performance improvement across successive generations.
    \item We demonstrate the capability of the proposed framework to customize suspension configurations for distinct driver behaviors and environments (e.g., mild driver vs. aggressive driver).
\end{itemize}
This work establishes a pathway toward more intelligent and adaptive DTs, offering enhanced ride comfort, robustness, and energy efficiency, and paving the way for personalized vehicle suspensions that evolve with driver behavior and environmental conditions.

The remainder of this article is organized as follows. {Section~\ref{sec2}} reviews the technical background on DTs, CCD formulations, and RL techniques. {Section~\ref{sec3}} formulates the full-vehicle active suspension dynamics, describes the numerical solution procedure, models external disturbances, and states the problem definition. 
{Section~\ref{sec4}} details the multi-generation CCD framework with the integration of DT and DRL, and its stepwise implementation. 
{Section~\ref{sec5}} presents results and extended discussion. 
Finally, {Section~\ref{sec6}} concludes the paper and outlines directions for future research.

\begin{figure*}
    \centering
    \includegraphics[width=0.7\linewidth]{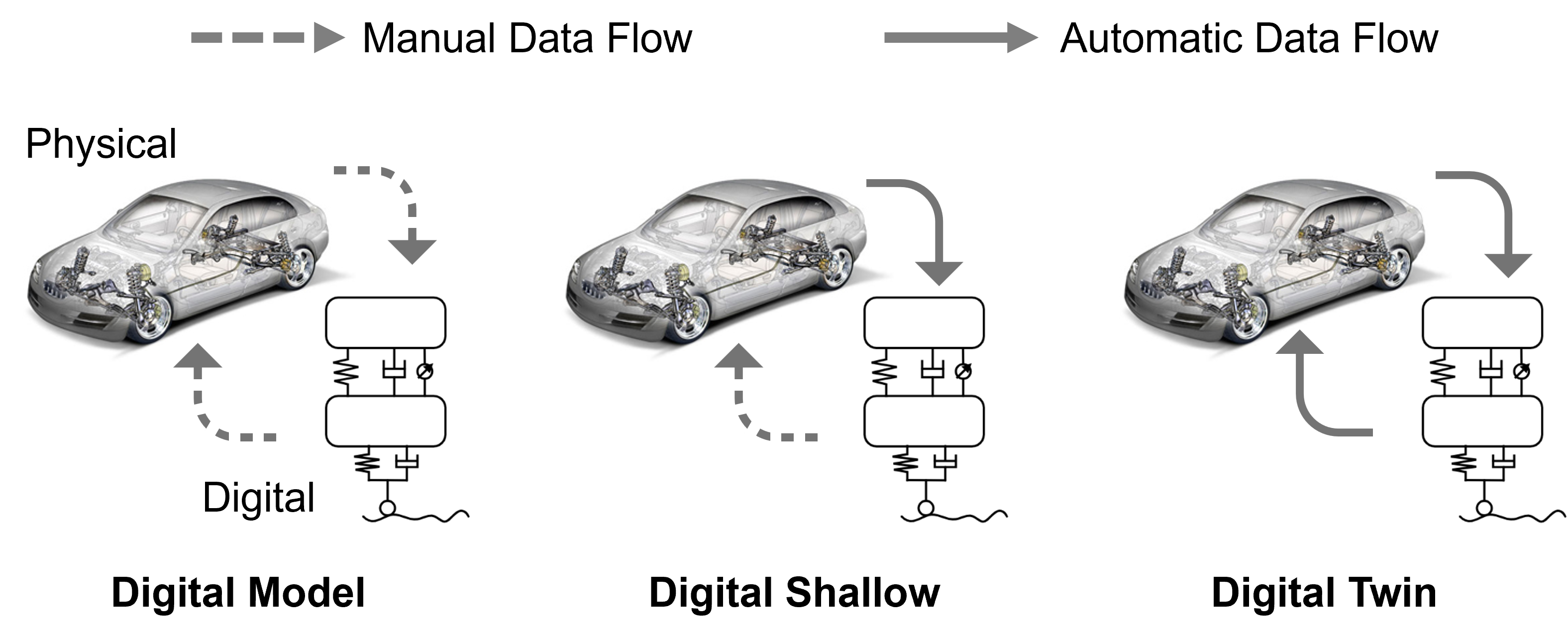}
    \caption{Illustration of the differences among a digital model, a digital shadow, and a digital twin.}
    \label{fig:DigitalModelShallowTwin}
\end{figure*}

\section{Background}\label{sec2}

\subsection{Digital Twin}
\label{subsec2_1}

A digital twin (DT) is generally defined as a virtual replica of a physical system that is continuously updated with data collected from the physical asset throughout its life cycle~\cite{grieves2014digital,thelen2022comprehensive,thelen2023comprehensive,karkaria2025optimization}. Unlike traditional modeling approaches that typically focus on a single design stage or a single generation of a product, a DT spans the entire product life cycle, from conceptual design~\cite{wang2020digital} and manufacturing~\cite{onaji2022digital} to operation~\cite{yu2021digital}, maintenance~\cite{karkaria2025digital,hu2024tutorial}, and next-generation redesign~\cite{van2023digital}. This continuous and bi-directional connection allows the DT not only to mirror the physical system but also to inform decision-making \cite{chen2025real} and enable adaptive improvements over time.

It is important to clarify the distinction between a digital model, a digital shadow, and a DT, as these terms are often used interchangeably in the literature~\cite{Grieves2017}. Figure~\ref{fig:DigitalModelShallowTwin} illustrates the three levels of digital representations using the example of an active suspension system. A digital model refers to a static virtual model of the physical system that is created during the design phase. This model is not automatically updated with real-world data. For example, a vehicle manufacturer may simulate suspension performance under pre-defined road profiles, but the model remains unchanged once it is built. A digital shadow, in contrast, is a virtual representation of the physical counterpart that receives a one-way flow of data from the physical system or real environment. Sensor measurements are streamed into the model, allowing it to reflect the current state of the physical system. However, the shadow does not send information back to the physical system. That is, no decision-support or feedback actions are transmitted from the digital to the physical domain. Finally, a DT represents a fully connected virtual replica of the physical system with bi-directional data flow. The twin not only updates its states with real-time sensor data but also uses predictive models and makes informed decisions for the real system. This closed-loop interaction enables the system to adapt to varying conditions and changing environments, thereby enhancing performance, robustness, and adaptability.

It is worth noting that in many practical cases, the bi-directional communication within a DT may not be fully automatic but instead involve human decision-makers in the loop~\cite{thelen2022comprehensive,sisson2022digital,marykovskiy2024architecting}. For example, predictive maintenance applications often rely on the DT to continuously monitor sensor data, identify early signs of component degradation, and recommend maintenance actions~\cite{karkaria2025digital}. Although the final decision and execution may require human approval, these actions must be handled in a timely manner to prevent unexpected breakdowns and ensure system reliability. Similarly, in remanufacturing scenarios, a DT can assess the condition of returned components, estimate their remaining useful life, and suggest refurbishment strategies, with human operators executing the recommended actions~\cite{karkaria2025ai}. In such cases, the DT functions as an intelligent decision-support system, ensuring that feedback between the physical and digital entities is closed effectively even when human involvement is necessary.


\subsection{Control Co-Design}
\label{subsec2_2}
Control Co-Design (CCD) is an integrated framework that concurrently addresses the design of physical systems (plants) and their control strategies, accounting for the interdependence between system dynamics and control behavior. Unlike conventional sequential approaches, CCD enables designers to achieve superior system-level performance by jointly considering both domains~\cite{garcia2019control,allison2014special,tsai2022constraint,van2022co}. Two prevalent CCD paradigms exist: simultaneous and nested formulations~\cite{herber2019nested}. The simultaneous approach treats physical and control variables within a unified optimization problem, whereas the nested approach decomposes the problem hierarchically, with the outer loop handling physical design and the inner loop solving for control under fixed physical parameters. The nested strategy is particularly advantageous when distinct objectives govern the control and physical design, or when the control synthesis relies on domain-specific methods~\cite{nash2021robust,tsai2023robust,tsai2025control}.

The choice between these formulations is guided by factors such as problem structure, computational considerations, and the tools available for controller synthesis. In this study, we adopt a simultaneous CCD framework, which enables co-optimization of the physical components and control policy through Deep Reinforcement Learning (DRL) with the integration of automatic differentiation. This learning-based method captures the complex interdependencies between design and control and facilitates system adaptability under dynamic and uncertain conditions by leveraging real-time data.

\subsection{Reinforcement Learning}
\label{subsec2_3}

\begin{figure}
    \centering
    \includegraphics[width=0.8\linewidth]{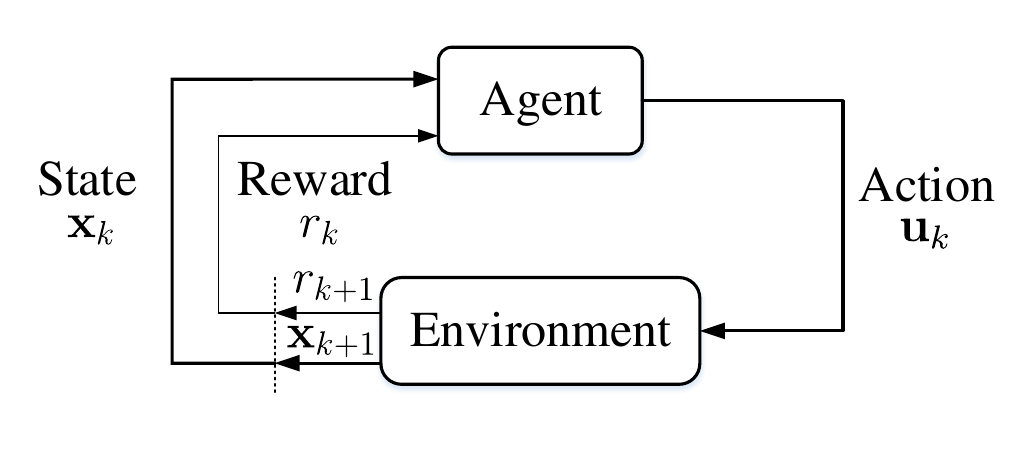}
    \caption{Diagram of reinforcement learning (RL), modified from \cite{sutton1998reinforcement}.}
    \label{fig:RL_diagram}
\end{figure}

Reinforcement learning (RL) provides a mathematical framework for solving sequential decision-making problems, where an agent learns to make decisions through interactions with its environment in order to maximize long-term rewards~\cite{bucsoniu2018reinforcement}, as illustrated in Fig.~\ref{fig:RL_diagram}. This learning process is typically modeled using a Markov Decision Process (MDP), represented by the tuple $(\mathcal{X},\mathcal{U},P,R,\gamma)$. Here, $\mathcal{X}$ denotes the state space, $\mathcal{U}$ the action (or control input) space, $P$ the system’s transition model, $R$ the reward function, which is often interpreted as the negative of a cost, and $\gamma\in(0,1]$ the discount factor. In an MDP framework, state transitions are generally stochastic, meaning that applying an action $\mathbf{u}_k$ to a system in state $\mathbf{x}_k$ at time step $k$ results in a probabilistic next state $\mathbf{x}_{k+1}$. The transition dynamics are captured by the probability distribution $P(\mathbf{x}_{k+1}|\mathbf{x}_k,\mathbf{u}_k)$, describing the likelihood of moving to state $\mathbf{x}_{k+1}$ from $\mathbf{x}_k$ under action $\mathbf{u}_k$. In control-oriented contexts, however, it is often more practical to express the system’s evolution deterministically as a function of state and control action:
\begin{equation}
    \mathbf{x}_{k+1}=\mathbf{f}(\mathbf{x}_k,\mathbf{u}_k),
\end{equation}
which is a stochastic process according to the transition function $P(\mathbf{x}_{k+1}|\mathbf{x}_k,\mathbf{u}_k)$.

After transitioning to the next state $\mathbf{x}_{k+1}$, the agent receives a scalar reward defined as $r_{k+1} = R(\mathbf{x}_k, \mathbf{u}_k, \mathbf{x}_{k+1})$, where the reward function $R: \mathcal{X} \times \mathcal{U} \times \mathcal{X} \rightarrow \mathbb{R}$ maps the current state, action, and resulting state to a real-valued reward. The agent's behavior is governed by a policy, which is a probabilistic mapping from states to actions. Specifically, given a current state $\mathbf{x}_k$, the policy $\boldsymbol{\pi}$ assigns probabilities to possible actions such that the agent selects $\mathbf{u}_k$ according to the distribution $\boldsymbol{\pi}(\mathbf{u}_k | \mathbf{x}_k)$.

The objective in RL is to maximize the return, defined as the cumulative discounted sum of future rewards starting from time step $k$:
\begin{equation}
G_k = \sum_{i=0}^{\infty} \gamma^i r_{k+i+1},
\end{equation}
where $\gamma \in (0, 1]$ is the discount factor, which determines the importance of future rewards relative to immediate ones.

To evaluate how good a state is under a given policy $\boldsymbol{\pi}$, we define the state-value function:
\begin{align}
V^{\boldsymbol\pi}(\mathbf{x}) &= \mathbb{E}_{\boldsymbol\pi}\left[G_k \mid \mathbf{x}_k = \mathbf{x} \right] \\\notag
&= \mathbb{E}_{\boldsymbol\pi} \left[ \sum_{i=0}^{\infty} \gamma^i R(\mathbf{x}_{k+i}, \mathbf{u}_{k+i}, \mathbf{x}_{k+i+1}) \mid \mathbf{x}_k = \mathbf{x} \right],
\end{align}
for all $\mathbf{x} \in \mathcal{X}$. This function, also called the V-function, estimates the expected return when starting from state $\mathbf{x}$ and following the policy $\boldsymbol{\pi}$ thereafter.

While the state-value function captures the quality of states under a policy, effective decision-making requires evaluating the quality of specific actions taken from a given state. This leads to the definition of the action-value function, or Q-function, as:
\begin{align}
&Q^{\boldsymbol\pi}(\mathbf{x}, \mathbf{u}) = \mathbb{E}_{\boldsymbol\pi} \left[G_k \mid \mathbf{x}_k = \mathbf{x}, \mathbf{u}_k = \mathbf{u} \right] \\\notag
&= \mathbb{E}_{\boldsymbol\pi} \left[ \sum_{i=0}^{\infty} \gamma^i R(\mathbf{x}_{k+i}, \mathbf{u}_{k+i}, \mathbf{x}_{k+i+1}) \mid \mathbf{x}_k = \mathbf{x}, \mathbf{u}_k = \mathbf{u} \right],
\end{align}
for all $\mathbf{x} \in \mathcal{X}$ and $\mathbf{u} \in \mathcal{U}$. The Q-function estimates the expected return of executing action $\mathbf{u}$ in state $\mathbf{x}$ and then continuing with policy $\boldsymbol{\pi}$ thereafter.

Conventional RL approaches, such as tabular Q-learning and dynamic programming, face significant challenges when dealing with complex control tasks involving continuous state and action spaces. In these scenarios, it is infeasible to explicitly represent or update the value function for all possible state-action combinations, as the space becomes effectively infinite. This scalability issue renders traditional RL methods unsuitable for high-dimensional control applications, including robotic manipulation, autonomous vehicle control, and real-time decision-making in DT-enabled systems.

\begin{figure*}
    \centering
	\begin{subfigure}{\textwidth}
        \centering
    	\includegraphics[width=0.6\textwidth]{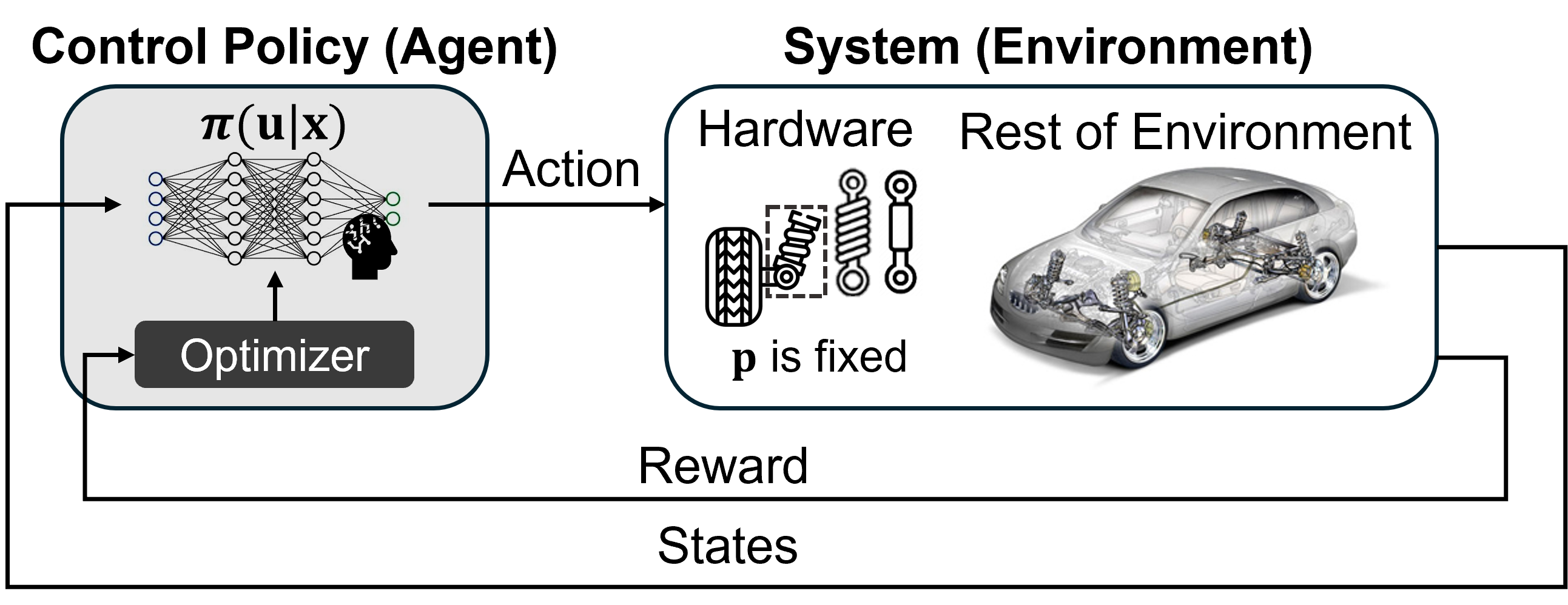}
    	\caption{}
    	\label{fig:TradControl}
	\end{subfigure}
    \hfill
	\begin{subfigure}{\textwidth}
        \centering
    	\includegraphics[width=0.6\textwidth]{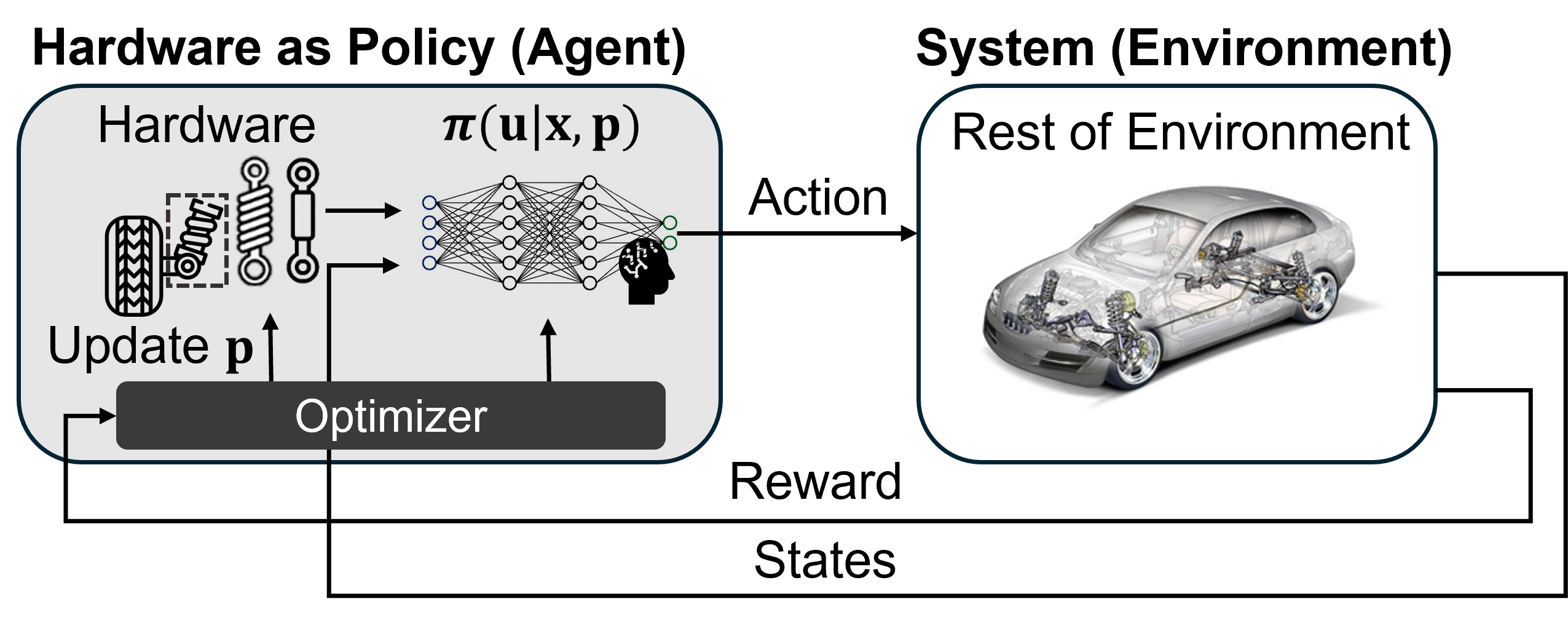}
    	\caption{}
    	\label{fig:Hardware_as_policy}
	\end{subfigure}
    \caption{Comparison of (a) conventional DRL-based control with fixed hardware parameter $\mathbf{p}$, where the policy $\boldsymbol\pi$ is only dependent of $\mathbf{p}$, and (b) hardware as policy, where the policy $\boldsymbol\pi$ is dependent of $\mathbf{x}$ and $\mathbf{p}$ and the hardware parameter and the policy are co-optimized (modified from \cite{chen2020hardware}).}
    \label{fig:tradcontrol_hardware_as_policy}
\end{figure*}

Deep reinforcement learning (DRL) overcomes the limitations of traditional RL methods in continuous and high-dimensional spaces by employing deep neural networks (DNNs) to approximate value functions, policies, or both~\cite{bucsoniu2018reinforcement,shakya2023reinforcement}. Rather than relying on explicit tables to store state-action values, DRL leverages the generalization capability of neural networks to estimate functions over large or continuous domains. This approach allows DRL to scale effectively to complex environments. Among the popular DRL algorithms are policy-based methods such as Trust Region Policy Optimization (TRPO)\cite{schulman2015trust} and Proximal Policy Optimization (PPO)\cite{schulman2017proximal}, which improve learning stability and efficiency through constrained updates. These methods ensure that policy changes remain within a “trust region,” thereby avoiding abrupt shifts that could hinder convergence. Beyond these, numerous DRL algorithms have been developed to address challenges like sample efficiency, exploration-exploitation trade-offs, and training stability. For broader overviews of the DRL landscape, readers may consult comprehensive surveys in~\cite{arulkumaran2017deep,ladosz2022exploration,wang2022deep}. The integration of deep learning into RL has dramatically expanded its practical relevance, enabling application to real-world systems where classical approaches fall short due to computational constraints.

\subsection{Co-Design using Reinforcement Learning}
\label{subsec2_4}
Given the remarkable success of DRL in solving complex control problems, its integration into co-design frameworks has attracted significant attention in recent years. In DRL-based co-design, the goal is to jointly optimize both the physical parameters of a system and its control policy, enabling a unified treatment of structural design and decision-making under uncertainty \cite{ma2021diffaqua,yuhn20234d,chen2020hardware,sun2023co,schaff2019jointly,luck2020data,wang2023preco}. This paradigm reflects a shift from the traditional sequential design–then–control process toward an adaptive, closed-loop optimization that allows physical morphology and control behavior to evolve together. Such co-adaptive mechanisms have shown improved robustness, adaptability, and performance in dynamic environments, particularly in robotic applications such as manipulators, legged locomotion, and modular soft robots.

A common strategy for solving learning-based co-design problems is to employ a bi-level optimization architecture, in which the inner loop trains a control policy for a fixed physical configuration, while the outer loop searches the design space through methods such as evolutionary algorithms or Bayesian optimization \cite{schaff2019jointly,wang2018neural,gupta2021embodied,chen2023c}. This separation allows each level of problem to use specific solvers: RL for control and black-box search for system design which is suitable for discrete or non-differentiable hardware spaces. However, the approach is often computationally intensive because each modification to the physical design requires retraining the controller from scratch, resulting in high sample cost and slow convergence \cite{chen2023evolving}.

To improve learning efficiency, an emerging line of research treats the mechanical design as part of the policy network itself, a concept known as hardware-as-policy \cite{chen2020hardware,sun2023co}. In this formulation, physical parameters are represented as differentiable components within the computational graph, allowing gradient-based DRL algorithms (e.g., PPO or TRPO) to co-optimize both neural-network weights and design variables through backpropagation \cite{schaff2019jointly,luck2020data,ma2021diffaqua}. Figure~\ref{fig:tradcontrol_hardware_as_policy} illustrates this concept by contrasting (a) conventional DRL-based control, where the hardware parameters $\mathbf{p}$ are fixed and the policy $\pi(\mathbf{u}|\mathbf{x})$ is learned independently, with (b) the hardware-as-policy paradigm, in which both $\mathbf{p}$ and the policy $\pi(\mathbf{u}|\mathbf{x},\mathbf{p})$ are jointly optimized through a shared computational graph. This simultaneous optimization significantly improves sample efficiency and coordination between morphology and control, producing systems that exhibit natural co-adaptation analogous to biological evolution. Despite their success, these methods typically assume stationary training environments and do not explicitly account for the stochastic, nonstationary conditions common in real engineering systems.

To address this gap, recent efforts have extended DRL-based co-design frameworks to engineering domains beyond robotics, where the physical dynamics are more complex, the environment is less predictable, and data collection is costly~\cite{he2023co,xu2017fast}. Our previous work \cite{tsai2025digital} represents one of the earliest applications of DRL-based CCD beyond robotics by demonstrating its use in a quarter-car active suspension system. That study introduced a gradient-based CCD formulation in which the physical suspension parameters and control policy were jointly optimized via automatic differentiation, enabling adaptive learning from road disturbances. Building on that foundation, the present study generalizes the framework to a full-vehicle active suspension system to facilitate real-time DT model and DRL policy update, co-optimize the physical system and controller, and improve the performance and adaptability across multiple generations.


\section{Simulation of Full Vehicle Active Suspension}\label{sec3}

Active suspension systems, unlike passive suspension systems, significantly enhance ride comfort and vehicle stability by actively modulating suspension forces in real time. Designing such systems demands a close integration of physical and control domains to ensure robust performance across diverse road and driving conditions. While traditional proportional–integral–derivative (PID) controllers and linear-quadratic regulators are widely used due to their simplicity~\cite{anh2020control,nguyen2023dynamic,manna2022ant}, they lack adaptability and require meticulous manual tuning, limiting their effectiveness in dynamic environments. While our prior work demonstrated that CCD of an active suspension system across multiple generations leads to improved performance and robustness~\cite{tsai2025digital}, it was limited to a quarter-car suspension model, which oversimplifies the complexity of real-world vehicle dynamics.

This section demonstrates the modeling and simulation of the full-vehicle active suspension system used in the case study. The objective is to illustrate how the vehicle dynamics are formulated and simulated under realistic operating conditions. The full-vehicle model serves as a digital representation that interacts with the DRL-based controller. We will use the simulation environment to show how the integration of real-time data collection and DT model updates enhances the system’s adaptability and dynamic performance (including driving quality, vehicle stability, and passenger comfort). Additionally, we will use the simulation environment to illustrate how the framework identifies distinct optimal solutions tailored to two distinct driving behaviors (mild and aggressive drivers) for personalized optimization and improving system robustness in real world scenarios.

\begin{figure}[t]
    \centering
    \includegraphics[width=\linewidth]{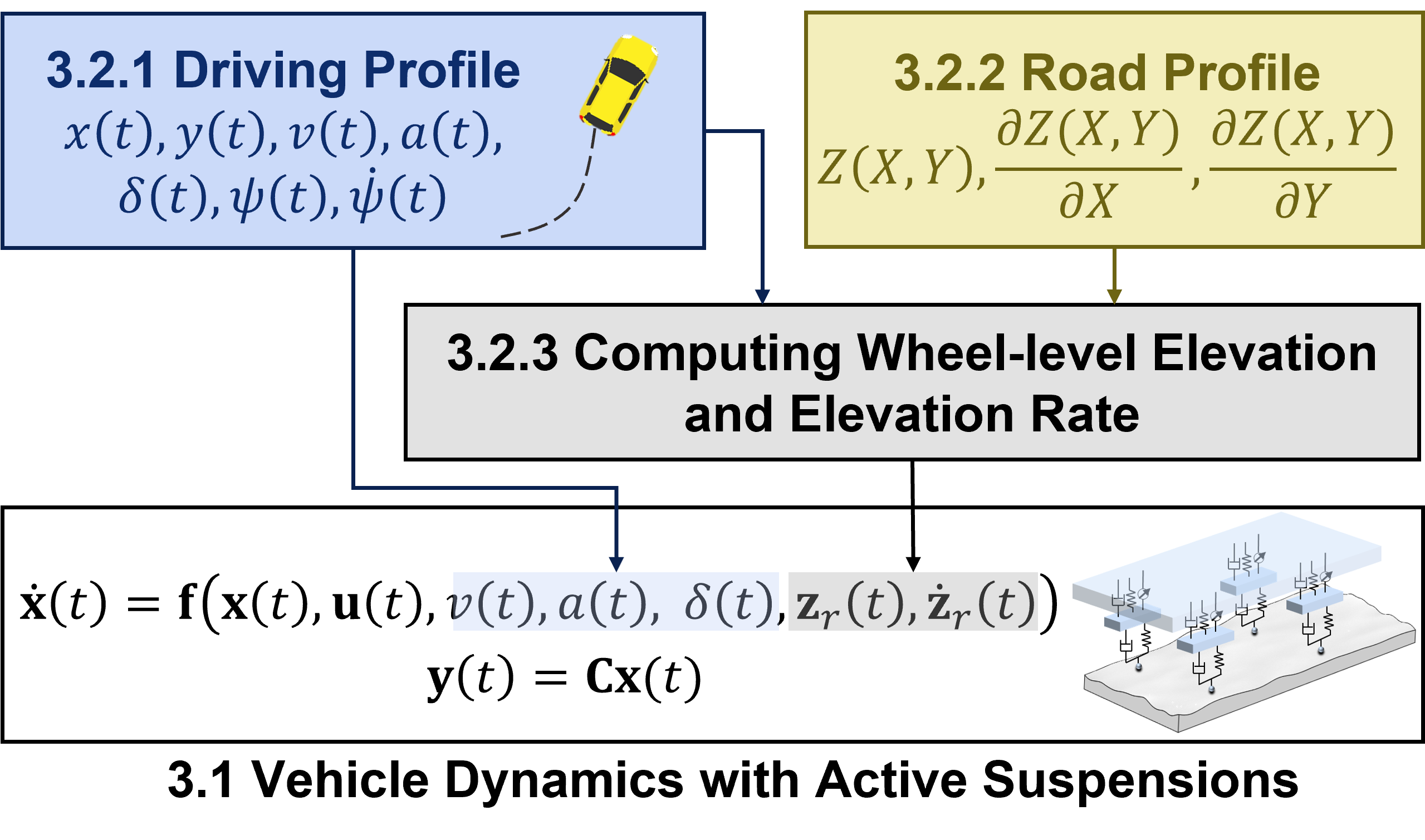}
    \caption{Simulation workflow for the full-vehicle active suspension system. The driving and road profiles are combined to compute wheel-level elevations and elevation rates, which serve as external disturbances to the dynamic model.}
    \label{fig:Simulation_FullVehcile_ActiveSuspension}
\end{figure}

Figure~\ref{fig:Simulation_FullVehcile_ActiveSuspension} illustrates the overall simulation flow for the full-vehicle active suspension system. The simulation begins by defining the {driving profile}, which specifies the vehicle’s longitudinal and lateral motion, including position, speed, acceleration, steering angle, and yaw dynamics. In parallel, the {road profile} describes the terrain elevation and its spatial gradients, capturing variations in road roughness and slope. These two components are combined to compute the {wheel-level elevation and elevation rate}, which act as external disturbances to the suspension system. The computed disturbances, together with the vehicle’s dynamic states and control inputs, form the inputs to the nonlinear state-space model of the vehicle dynamics. The model then predicts the time evolution of the system states and observable outputs, providing a realistic representation of full-vehicle behavior under diverse driving and road conditions.

\subsection{Dynamic Modeling}
\label{subsec3_1}

The vehicle dynamics with active suspension systems are represented as a nonlinear state-space system:
\begin{equation}
    \dot{\mathbf{x}}(t)=\mathbf{f}\left(\mathbf{x}(t),\mathbf{u}(t),v(t),a(t),\delta(t),\mathbf{z}_r(t),\dot{\mathbf{z}}_r(t)\right),
    \label{eq:continuous_dynamics}
\end{equation}
where $\mathbf{x}(t)$ is the system state vector, defined by:
\begin{equation}
    \begin{split}
        \mathbf{x}(t)=\left[z_s(t),\alpha(t),\beta(t),z_{u1}(t),z_{u2}(t),z_{u3}(t),z_{u4}(t),\right.\\
        \left.\dot{z}_s(t),\dot{\alpha}(t),\dot{\beta}(t),\dot{z}_{u1}(t),\dot{z}_{u2}(t),\dot{z}_{u3}(t),\dot{z}_{u4}(t)\right]^\top,
    \end{split}  
\end{equation}
$\mathbf{u}(t)=[u_1(t),u_2(t),u_3(t),u_4(t)]^\top$ is the vector of actuator control forces applied to the four suspension systems, $v(t)$ and $a(t)$ denote the vehicle's longitudinal velocity and acceleration, $\delta(t)$ is the steering angle, and $\mathbf{z}_r(t)=[z_{r1}(t),z_{r2}(t),z_{r3}(t),z_{r4}(t)]^\top,\dot{\mathbf{z}}_r(t)=[\dot{z}_{r1}(t),\dot{z}_{r2}(t),\dot{z}_{r3}(t),\dot{z}_{r4}(t)]^\top$ are the vectors of road heights and road velocity inputs at four wheels. 

The dynamic equations of sprung mass vertical motion, pitch motion, roll motion, and unsprung mass (wheel) vertical motion, respectively, are:
\begin{align}
m_s \ddot{z}_s(t) &= \sum_{i=1}^{4} \left( F_{si}(t) + u_i(t) \right), \\
I_\alpha(t) \ddot{\alpha} &= \sum_{i=1}^{4} \left( F_{si}(t) + u_i(t) \right) d_{xi} + M_\alpha(t),\\
I_\beta(t) \ddot{\beta} &= \sum_{i=1}^{4} \left( F_{si}(t) + u_i(t) \right) d_{yi} + M_\beta(t),\\
m_{ui}\ddot{z}_{ui}(t) &= F_{ti}(t)-F_{si}(t)-u_i(t),~\forall i\in\{1,2,3,4\},
\end{align}
where $F_{si}(t)$ denotes the suspension force at Wheel $i$, defined by:
\begin{equation}
    F_{si}(t) = k_s (z_{ui}(t) - z_s(t) - \Delta_i(t)) + c_s (\dot{z}_{ui}(t) - \dot{z}_s(t) - \dot{\Delta}_i(t)),
\end{equation}
and $F_{ti}(t)$ denotes the tire (contact) force at Wheel $i$, defined by:
\begin{equation}
    F_{ti}(t) = k_t (z_{ri}(t) - z_{ui}(t)) + c_t (\dot{z}_{ri}(t) - \dot{z}_{ui}(t)),
\end{equation}
where $\Delta_i$ accounts for the geometric displacement due to pitch and roll at each wheel location:
\begin{align}
    \Delta_1(t)&=-l_f\alpha(t)+\frac{l}{2}\beta(t),\\
    \Delta_2(t)&=-l_f\alpha(t)-\frac{l}{2}\beta(t),\\
    \Delta_3(t)&=l_f\alpha(t)+\frac{l}{2}\beta(t),\\
    \Delta_4(t)&=l_f\alpha(t)-\frac{l}{2}\beta(t),
\end{align}
$d_{xi}=\{-l_f,-l_f,l_r,l_r\}$ and $d_{yi}=\{l/2,-l/2,l/2,-l/2\}$ denote the longitudinal and lateral distances, respectively, from the vehicle’s center of gravity to each wheel, $k_t$ and $c_t$ are tire stiffness and damping constant, respectively, and $k_s$ and $c_s$ are the coefficients of the spring and the damper (design variables for the physical system). The additional torques $M_\alpha(t)$ and $M_\beta(t)$ are generated by inertial coupling due to longitudinal and lateral accelerations (e.g., braking and cornering), modeled as $M_\alpha(t)=m_sh_{CG}a(t)$ and $M_\alpha(t)=m_sh_{CG}\frac{v(t)^2\tan\delta(t)}{l_f+l_r}$, with $h_{CG}$ representing the height of the vehicle’s center of gravity (CG) above ground. The physical and geometric parameters used in the full-vehicle dynamic model are summarized in Table~\ref{tab:vehicle_para}.

\begin{figure}
    \centering
    \includegraphics[width=0.85\linewidth]{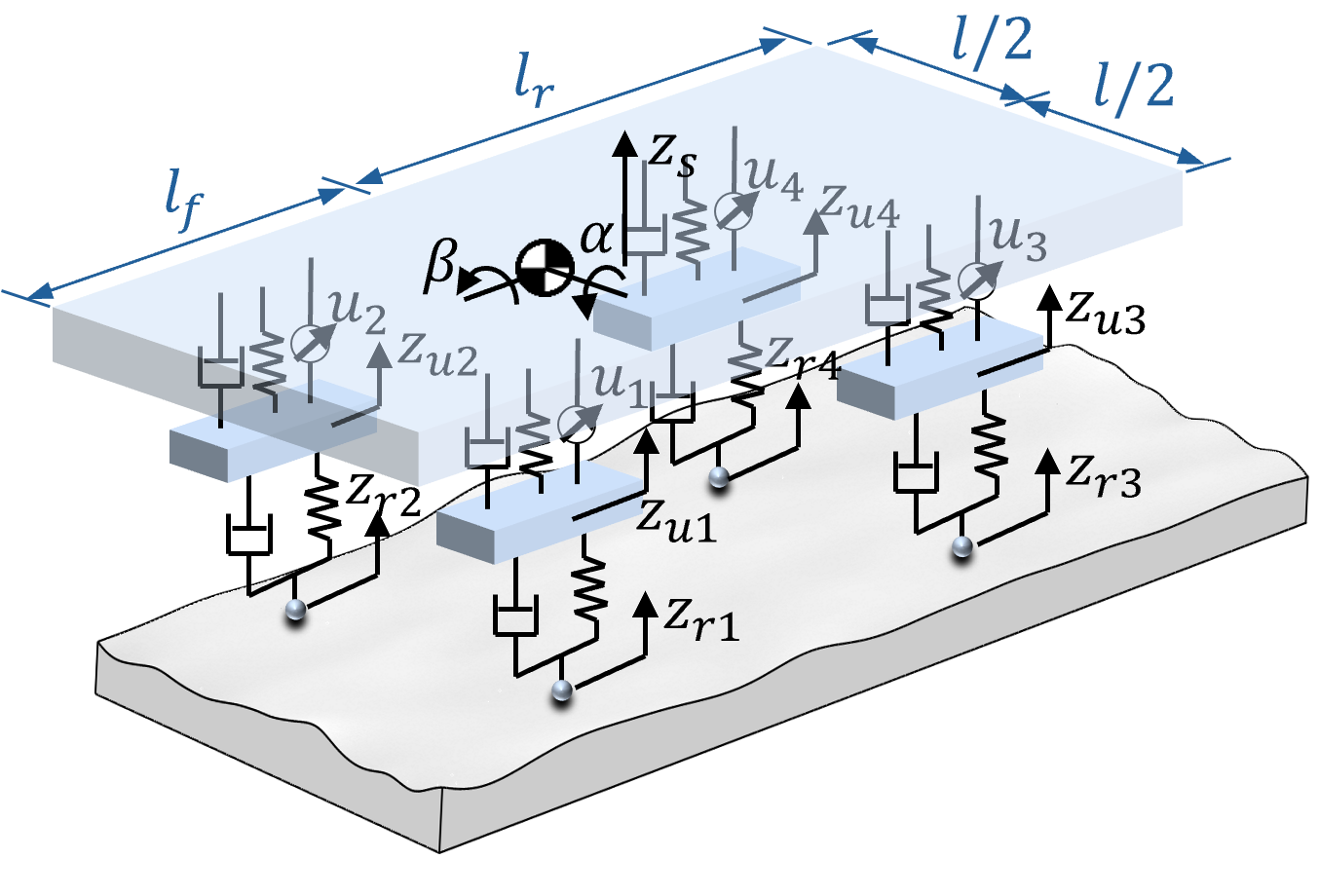}
    \caption{Model of full vehicle with active suspension systems.}
    \label{fig:FullVehicleActiveSuspension}
\end{figure}

This model captures seven degrees of freedom (DOFs): three DOFs for the sprung mass (lift $z_s$, pitch $\alpha$, and roll $\beta$), and four vertical DOFs for the unsprung masses (wheels) $z_{u1}-z_{u4}$, shown in Fig.~\ref{fig:FullVehicleActiveSuspension}. We assume small angles for pitch and roll to simplify the rotational dynamics. Tire-road interactions are modeled with vertical spring-damper pairs, and control forces $\mathbf{u}(t)=[u_1(t),u_2(t),u_3(t),u_4(t)]^\top$ are treated as external actuator inputs that directly adjust the actuation forces for the four suspension systems.

\begin{table*}
\small
\centering
\caption{Vehicle parameters.}
    \begin{tabular}{llcc}
    \hline
         Symbol & Physical Meaning & Value & Unit  \\\hline
         $m_s$ & Sprung mass & 1500 & kg\\\hline
         $I_\alpha$ & Pitch rotation inertia & 2500 & kg m${}^{2}$\\\hline
         $I_\beta$ & Roll rotation inertia & 500 & kg m${}^{2}$\\\hline
         $m_{us}$ & Unsprung mass & 50 & kg\\\hline
         $k_t$ & Stiffness of tire & 200000 & N/m\\\hline
         $c_t$ & Damping of tire & 150 & N s/m\\\hline
         $l_f$ & Distance between CG and front axles & 1.35 & m\\\hline
         $l_r$ & Distance between CG and rear axles & 1.35 & m\\\hline
         $l$ & Track front, track rear & 0.75 & m\\\hline
         $h_{CG}$ & Height of the CG above the ground & 0.55 & m\\\hline
    \end{tabular}
    \label{tab:vehicle_para}
\end{table*}

Suspension systems often suffer from limited observability due to the difficulty of directly measuring all internal states in real-time, especially the motion and dynamics of the sprung body. In particular, while the full vehicle dynamics involve 14 states, only a subset can be observed through sensors. In this system, the observed state vector $\mathbf{y}(t)$ is defined by:
\begin{equation}
\begin{split}
    \mathbf{y}(t)=[\dot{z}_s(t),\dot{\alpha}(t),\dot{\beta}(t),\\z_{u1}(t),z_{u2}(t),z_{u3}(t),z_{u4}(t),\\
        z_s(t) - z_{u1}(t),z_s(t) - z_{u2}(t),\\z_s(t) - z_{u3}(t),z_s(t) - z_{u4}(t)]^\top\\=\mathbf{C}\mathbf{x}(t),
\end{split}
\end{equation}
where $\mathbf{C}\in\mathbb{R}^{11 \times 14}$ is the observation matrix, allowing us to extract the following 11 measurable outputs from the full system. The vertical velocity of the vehicle body \( \dot{z}_s(t) \), and pitch rate \( \dot{\alpha}(t) \) and roll rate \( \dot{\beta}(t) \) can be obtained from a gyroscope module. The vertical positions of the unsprung masses \( z_{ui}(t) \), \( i = 1,\dots,4 \), are measured via linear potentiometers attached to each wheel. Additionally, the relative displacement between the body and wheels \( z_s(t) - z_{ui}(t) \) is measured using suspension stroke sensors.

To implement the dynamic model into the DRL framework where the agent interacts with the environment at discrete time intervals, the continuous-time vehicle dynamics are discretized using the fourth-order Runge–Kutta (RK4) method. Given the current system state $\mathbf{x}(t)$ and control input $\mathbf{u}(t)$, the RK4 method computes the next state $\mathbf{x}_{k+1}$ over a time step $\Delta t=0.01$ second by evaluating intermediate derivatives of the dynamics function $\mathbf{f}$, where $k$ represents the index of the time step. With the prescribed $\Delta t$, the dynamics can be approximated and Equation (\ref{eq:continuous_dynamics}) can be written in a discrete-time version:
\begin{equation}
    \mathbf{x}_{k+1} = \mathbf{f_d}(\mathbf{x}_k, \mathbf{u}_k, v_k,a_k,\delta_k,\mathbf{z}_{rk},\dot{\mathbf{z}}_{rk}),
    \label{eq:discrete_dynamics}
\end{equation}
which serves as the transition model in the agent–environment interaction loop. It is noted that the full system state $\mathbf{x}_k$ is not directly observable in practice. 
The agent instead receives a partially observed state (or observation) vector $\mathbf{y}_k$. The DRL policy $\boldsymbol{\pi}(\mathbf{u}_k|\mathbf{y}_k)$ is therefore trained on the observation space rather than the full state, reflecting the realistic feedback structure in which the controller operates using only sensor-accessible information.

The vehicle dynamic model used in this study is based on several simplifying assumptions. First, it is assumed that all four wheels maintain continuous contact with the road surface, i.e., no wheel lift-off or loss of ground contact occurs. The road excitations are applied only in the vertical direction, and lateral or longitudinal tire slip dynamics are not considered. In addition, internal disturbances such as drivetrain vibrations, engine torque fluctuations, or transmission dynamics are neglected~\cite{dridi2025optimizing}. The suspension elements are assumed to be ideal, with no friction, backlash, or hysteresis effects, and the actuators apply control forces directly without delay or saturation. Sensor measurements are assumed to be noise-free and perfectly accurate.

This full-vehicle suspension model presents several challenges for control:
\begin{itemize}
    \item High dimensionality and coupled dynamics,
    \item Partial observability (not all states may be measurable in practice), i.e., only $\mathbf{y}_{k}$ is observable instead of $\mathbf{x}_{k}$ at time step $k$,
    \item Highly uncertain environments, such as uneven terrain and aggressive driving maneuvers, and
    \item The need for rapid and adaptive decision-making to maintain vehicle stability and passenger comfort.
\end{itemize}

DRL updates the control policy in real time by learning from physical data to adapt to dynamic environments. Furthermore, DRL is well-suited for systems where the cost of modeling uncertainty and real-time feedback is high, and where traditional control strategies may struggle to maintain performance under unpredictable driving conditions.

\subsection{External Disturbance}

\subsubsection{Driving Profiles}
\label{subsec3_2_1}
To demonstrate the capability of the proposed DT-enabled CCD framework in tailoring optimal suspension system designs for different user behaviors, we developed two contrasting driving profiles representative of mild and aggressive drivers. These profiles serve as external excitation scenarios, enabling evaluation of vehicle performance and adaptation under varied longitudinal and lateral dynamics.

The mild driver is characterized by gradual acceleration, moderate steering actions, and a lower cruising speed. In contrast, the aggressive driver exhibits rapid acceleration, more abrupt steering transitions, and a higher cruising speed. These behavioral patterns are embedded in the acceleration and steering angle trajectories, which serve as inputs for vehicle motion integration.

\begin{figure}[t]
\centering
\includegraphics[width=\linewidth]{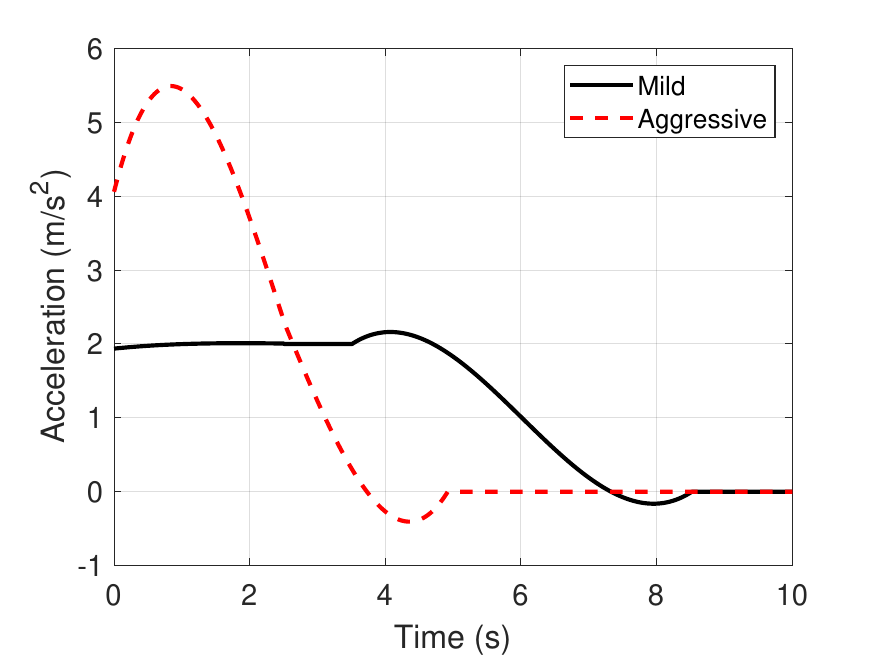}
\caption{Comparison of longitudinal acceleration profiles for mild and aggressive drivers. The aggressive driver accelerates more quickly and reaches a higher cruising speed.}
\label{fig:accel_profile}
\end{figure}

We constructed both profiles over a duration of {1,200} s with a step of {0.01} s (a total of 120,000 steps). The longitudinal acceleration profile $a(t)$ was divided into three phases: acceleration, cruising, and braking. The mild driver accelerates at {2.0} {m/s${}^2$} to a cruising speed of {12} {m/s}, while the aggressive driver reaches {20} {m/s} under a higher acceleration of {6.0} {m/s${}^2$} (shown in Fig.~\ref{fig:accel_profile}).

\begin{figure}[t]
\centering
\includegraphics[width=\linewidth]{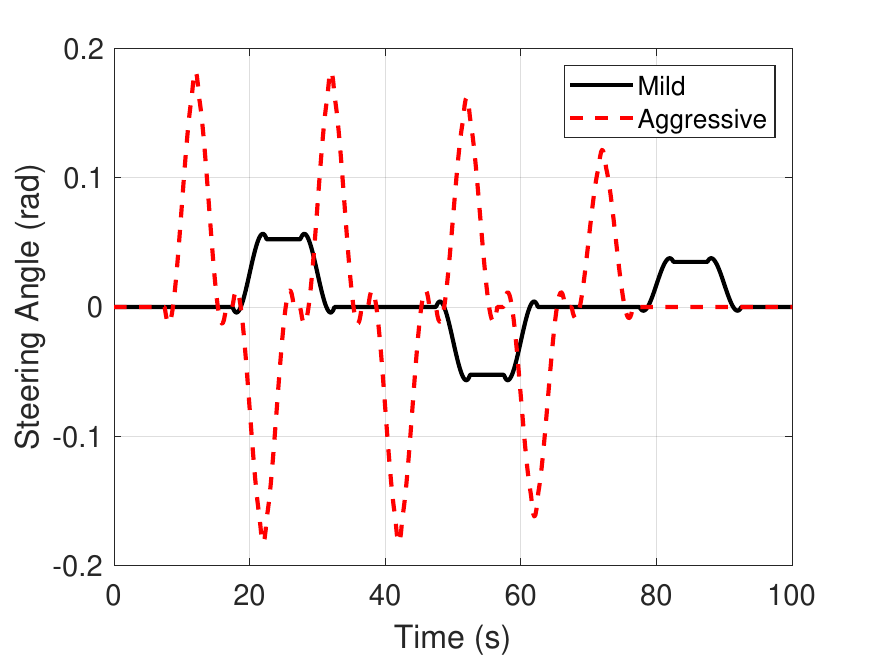}
\caption{Steering angle trajectories for mild and aggressive drivers. The aggressive driver demonstrates higher frequency and amplitude of steering commands.}
\label{fig:steering_profile}
\end{figure}

Lateral control is represented by scheduled steering angle inputs, modeled using piecewise-defined angular perturbations repeated over the full simulation. The aggressive driver performs frequent and sharp turns (up to approximately 0.17 rad), while the mild driver exhibits infrequent and smoother maneuvers (within around 0.05 rad), shown as Fig.~\ref{fig:steering_profile}.

All profiles were smoothed using a Savitzky-Golay filter~\cite{schmid2022and} to emulate realistic driver commands while preserving sharp dynamic features. The resulting position $(x,y)$, velocity $v$, yaw $\psi$, and yaw rate $\dot{\psi}$ define the vehicle’s trajectory across the road surface, from which the spatially varying road elevations are queried at each wheel as a function of time. This ensures that the excitation inputs applied to the four suspension systems correspond accurately to the instantaneous contact locations of the tires as the vehicle moves along the road, which will be detailed in Section \ref{sec3.2.3}.


Figure~\ref{fig:accel_profile} and Figure~\ref{fig:steering_profile} illustrate selected segments of the longitudinal acceleration and steering angle profiles, respectively, highlighting the key differences between the driving behaviors. The aggressive profile features pronounced peaks in acceleration and steering, leading to sharper velocity and orientation changes. The mild profile, in contrast, emphasizes smoothness and gradual transitions. These user-specific trajectories are used as input scenarios to evaluate the adaptability of the suspension design under different driving styles.

\subsubsection{Road Profile}
\label{subsec3_2_2}
To evaluate the performance of the full vehicle active suspension system under realistic conditions, we generated a high-fidelity two-dimensional (2D) road surface profile $Z(X, Y)$ that captures both micro-scale stochastic roughness and macro-scale topographic features. This approach goes beyond traditional one-dimensional bump or cleat models used in~\cite{dridi2025optimizing} by introducing lateral variability and a spatially continuous field suitable for evaluating vehicle responses across all four wheels simultaneously.


The 2D road roughness profile was synthesized using inverse Fourier transformation of a Hermitian-symmetric complex spectrum constructed according to the ISO~8608 standard. The power spectral density (PSD) was defined as:
\begin{equation}
    \Phi(n_X, n_Y) = S_0 \left( \frac{n_0}{\sqrt{n_X^2 + n_Y^2 + \epsilon}} \right)^\omega
    \label{eq:PSD}
\end{equation}
where $S_0$ is the reference PSD value (e.g., $1 \times 10^{-4} \, \text{m}^3$), $n_X$ and $n_Y$ represent the numbers of grids in x and y directions, respectively, \( n_0 = 0.1 \, \text{cycles/m}\) is the reference spatial frequency, and \( \omega = 2.5 \) is the waviness exponent. A small constant \( \epsilon=0.005 \) m was introduced to avoid division by zero. To ensure a real-valued elevation field, Hermitian symmetry was enforced on the frequency domain matrix prior to applying the inverse FFT.

\begin{figure}[t]
    \centering
    \includegraphics[width=1.05\linewidth]{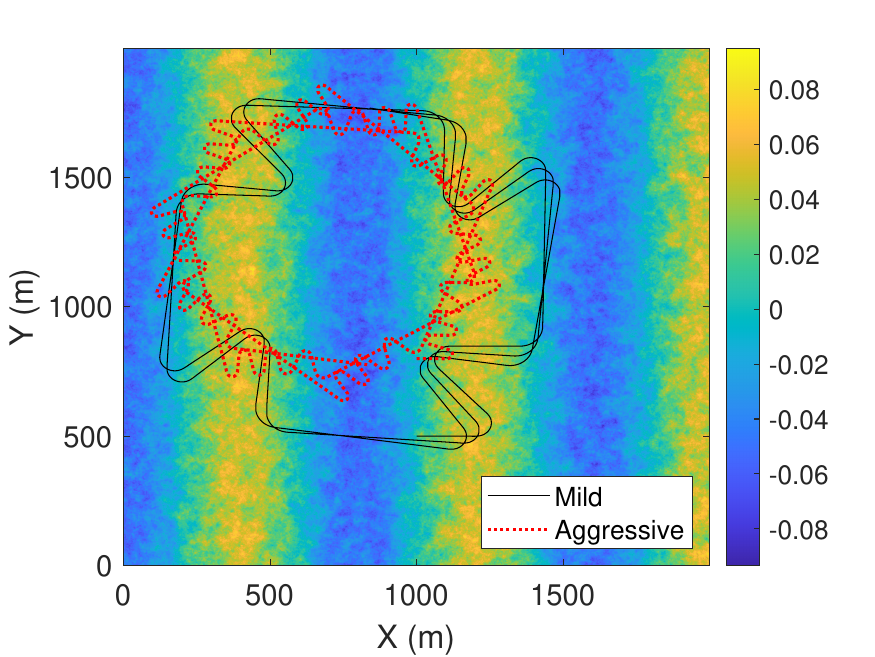}
    \caption{Generated 2D road roughness profile based on ISO~8608-based spectral characteristics, with paths of mild and aggressive drivers.}
    \label{fig:road_profile}
\end{figure}

The resulting elevation map \( Z(X, Y) \in \mathbb{R}^{N_X \times N_Y} \) spans a \( 2000\,\text{m} \times 2000\,\text{m} \) domain with 1-meter spatial resolution. An example of the generated roughness is shown in Fig.~\ref{fig:road_profile}, exhibiting spatial correlation with a standard deviation of approximately 0.045 m and peak-to-peak variation exceeding 0.15 m.


To simulate structured terrain features such as hills or ramps, we superimposed large-scale elevation patterns onto the stochastic roughness field. For instance, sinusoidal hills were defined in the longitudinal direction using:
\[
Z_{\text{hill}}(X) = A_{\text{hill}} \cdot \sin\left( \frac{\pi (X - X_0)}{L_{\text{hill}}} \right)
\]
where \( A_{\text{hill}} = {0.05}\text{ m} \) is the hill amplitude, \( X_0 = {1000}\text{ m} \) is the hill’s starting position, and \( L_{\text{hill}} = {400}\text{ m} \) is the hill length. This design introduces realistic undulations similar to those found in rural or suburban roads.


To couple the road profile with vehicle dynamics, we implemented a spatial interpolator using \texttt{scipy.interpolate.RectBivariateSpline}, which provides smooth querying of \( Z(X, Y) \) and its spatial gradients at arbitrary coordinates. The road elevation and partial derivatives \( \partial Z / \partial X \) and \( \partial Z / \partial Y \) were computed at each wheel based on the vehicle’s pose (position and yaw angle) and geometric configuration (wheelbase and track width). This allows direct integration of terrain excitation into the suspension dynamics model and the RL environment.

\subsubsection{Computing Wheel-Level Elevation and Elevation Rate}\label{sec3.2.3}

After generating the two-dimensional road profile \( Z(X, Y) \) and defining the vehicle motion trajectories, we compute the road excitation input at each suspension point in the form of elevation \( \mathbf{z}_r \) and elevation rate \( \dot{\mathbf{z}}_r \), which appear in Eq.~(\ref{eq:discrete_dynamics}) of the suspension dynamics model. These quantities serve as exogenous inputs to the vertical dynamics of each wheel.


At each simulation timestep, the global position of the vehicle’s center of gravity (CG), denoted as \( (x(t), y(t)) \), and its yaw angle \( \psi(t) \), are used to compute the global positions of the four suspension contact points:
\begin{itemize}
    \item Front left (FL): $(\Delta x,\Delta y)=(l_f,-l/2)$
    \item Front right (FR): $(\Delta x,\Delta y)=(l_f,l/2)$
    \item Rear left (RL): $(\Delta x,\Delta y)=(-l_f,-l/2)$
    \item Rear right (RR): $(\Delta x,\Delta y)=(-l_f,l/2)$
\end{itemize}

Each wheel’s location is determined by rotating its local offset relative to the CG using the yaw angle:
\[
\begin{aligned}
x_{\text{wheel}} &= x + \Delta x \cos\psi - \Delta y \sin\psi, \\
y_{\text{wheel}} &= y + \Delta x \sin\psi + \Delta y \cos\psi,
\end{aligned}
\]
where \( \Delta x \) and \( \Delta y \) are the relative longitudinal and lateral distances from the CG to each wheel, determined by $l_f$ and $l$.


Given the global position of each wheel, the corresponding road elevation \( {z}_{ri} \) is extracted from the spatial map \( Z(X, Y) \). In addition, the local terrain slopes \( \partial Z / \partial X \) and \( \partial Z / \partial Y \) are obtained from pre-computed gradient maps of the road surface. These values are queried using bilinear or spline-based interpolation to ensure smoothness.


The vertical rate of change of road elevation at each wheel contact point, denoted \( \dot{z}_{ri} \), is calculated using the chain rule:
\[
\dot{z}_{ri} = \frac{\partial Z_i}{\partial Y} v\cos\psi + \frac{\partial Z_i}{\partial Y} v\sin\psi,
\]
where \( v \) is the longitudinal velocity of the vehicle. This formulation allows the model to capture how the suspension experiences road gradients based on the direction and speed of vehicle motion.

\subsection{Problem Definition}
\label{subsec3_4}

The goal of active suspension control is to enhance both ride comfort and vehicle stability by actively adjusting the forces applied at each suspension. However, achieving optimal performance requires careful coordination between the physical design parameters (e.g., suspension stiffness and damping) and the control policy that governs the actuator behavior.

In our framework, we focus on co-designing the mechanical spring and damper parameters, i.e., $k_s$ and $c_s$, for each of the four suspension units. These components play a critical role in determining how effectively the suspension system can isolate the vehicle body from road disturbances while maintaining contact between the tires and the ground.

To evaluate the performance of each suspension-controller configuration, we define a reward function that reflects key objectives in ride comfort, vehicle handling, and control efficiency. The reward is computed as a weighted sum of physically meaningful performance indices derived from the dynamic responses. Specifically, ride comfort is quantified using a comfort index, which aggregates vertical acceleration, pitch acceleration, and roll acceleration, defined by:
\begin{equation}
    \text{comfort index}=\sqrt{(w_1\ddot{z}_{sk})^2+(w_2\ddot{\alpha}_k)^2+(w_3\ddot{\beta}_k)^2},
    \label{eq:comfort_index}
\end{equation}
where $w_1=10$, $w_2=1.0$, and $w_3=0.5$ are scaling weights to account for their relative contributions to passenger discomfort, with vertical acceleration being the dominant factor.

In addition to comfort, handling performance is evaluated through penalties on pitch and roll angles, which are squared to reflect their increasing impact on stability as the magnitude grows. These penalties are normalized using empirically chosen scaling factors to ensure appropriate weighting within the total cost. Additionally, the reward also includes an energy penalty proportional to the squared control inputs applied at each suspension. Altogether, the reward function balances the competing objectives of comfort, handling, and control effort:
\begin{equation}
    r_{k+1}=-\left(\text{comfort index}+c_1\alpha_k^2+c_2\beta_k^2+c_3\sum_{i=1}^4u_{ik}^2\right),
    \label{eq:reward_susp}
\end{equation}
where $c_1=\frac{1}{0.00004}$, $c_2=\frac{1}{0.00003}$, and $c_3=0.0001$ are coefficients according to the quantity scales.

\section{Multi-generation Digital Twin Framework}\label{sec4}

\subsection{Overview}
\label{subsec4_1}

\begin{figure}[t]
    \centering
    \includegraphics[width=0.65\linewidth]{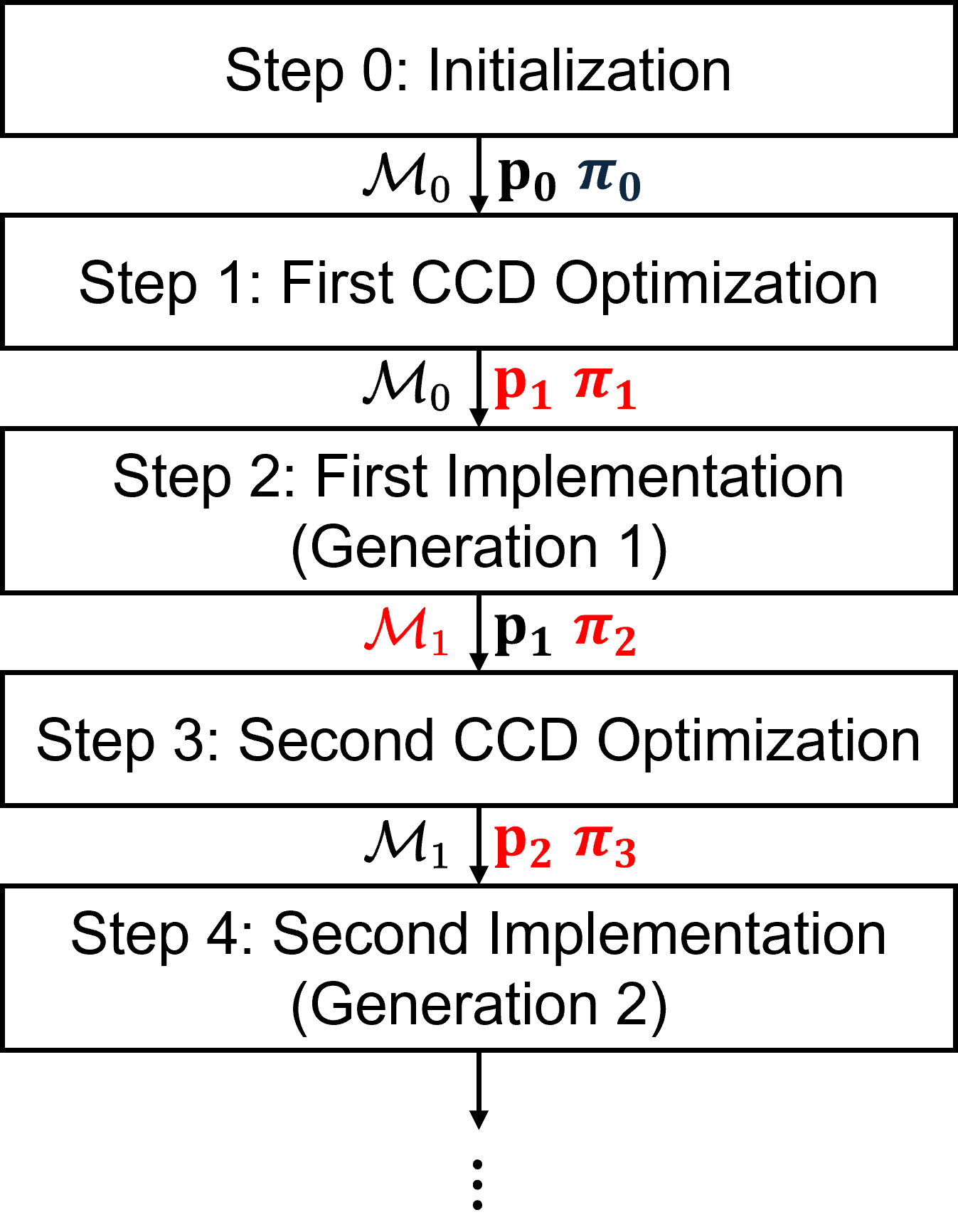}
    \caption{Overview of the multi-generation digital twin-based control co-design framework. Updated components at each stage are highlighted in red. ($\mathcal{M}$: digital model, $\mathbf{p}$: system design, and $\boldsymbol{\pi}$: controller)}
    \label{fig:MultigenerationCCD_flowchart}
\end{figure}

The framework proposed in~\cite{tsai2025digital} establishes a closed learning loop between the digital and physical domains, enabling continuous improvement of both system design and control policy across multiple generations. Building on this foundation, the present work applies and extends the framework to the optimization of a full-vehicle active suspension system, where the DT learns and adapts to distinct driving behaviors and varying environmental conditions. As illustrated in Fig.~\ref{fig:MultigenerationCCD_flowchart}, the process begins with an initial policy~$\boldsymbol{\pi}_0$ and baseline digital model~$\mathcal{M}_0$ (\textbf{Step 0}), followed by the first CCD optimization (\textbf{Step 1}) that jointly determines the physical design variables~$\mathbf{p}_1$ and the control policy~$\boldsymbol{\pi}_1$. These optimized configurations are then implemented on the physical platform (\textbf{Step 2}, Generation 1), where operational data are collected under realistic environmental and loading conditions.

\begin{figure*}[t]
    \centering
    \includegraphics[width=\linewidth]{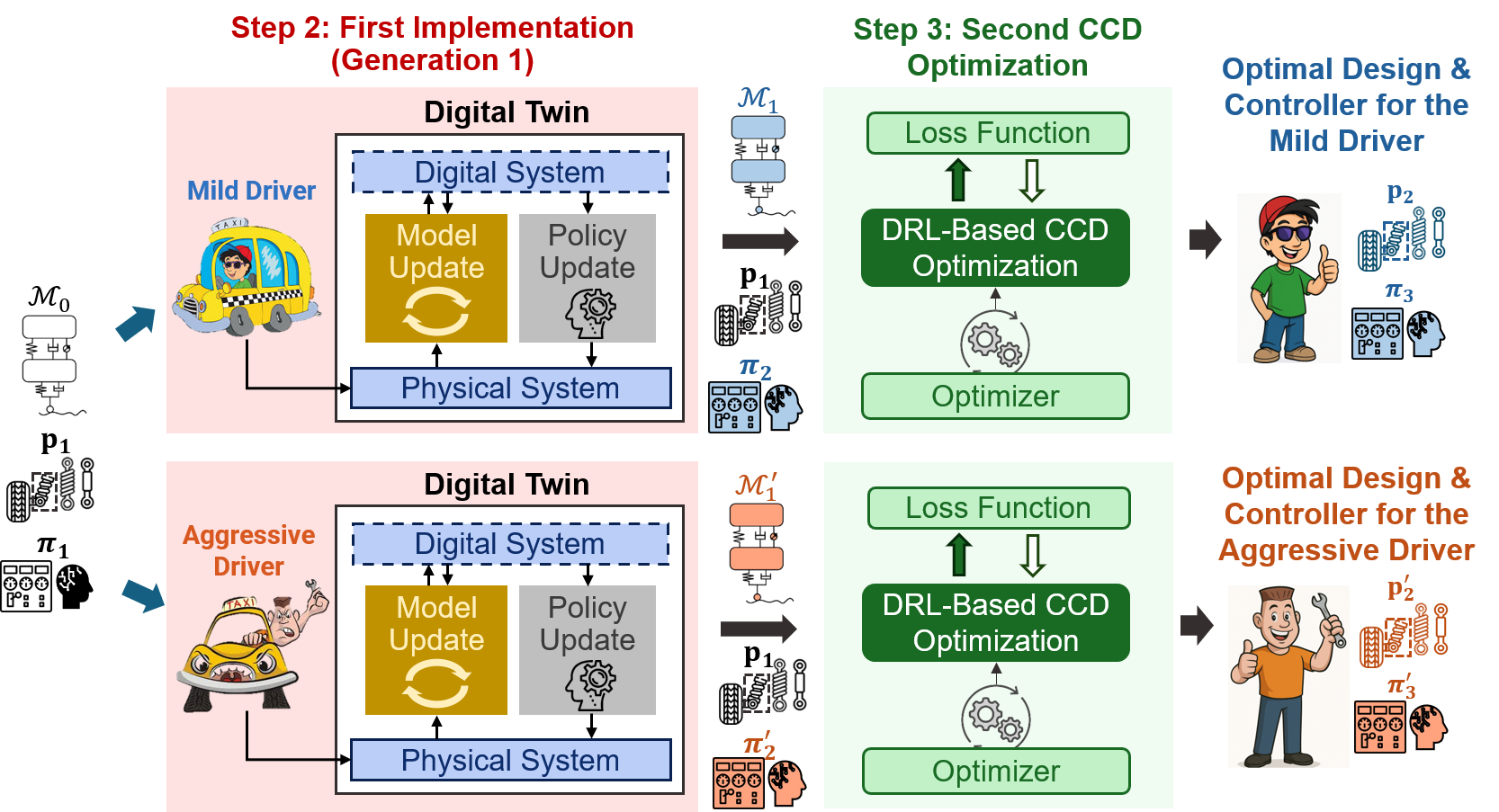}
    \caption{Customization of active suspension designs and controllers for distinct driving behaviors after \textbf{Step 1} first CCD optimization. Through data collection, real-time model and policy updates, and redesign with the updated models and policies, we can yield customized and optimal solutions for different drivers. The mild and aggressive drivers exhibit distinct behaviors that significantly influence suspension dynamics. By continuously updating the virtual systems (from $\mathcal{M}_0\to\mathcal{M}_1$ and $\mathcal{M}_1'$) and control policies based on physical data, and re-optimizing the design-controller pair, the resulting solutions $(\mathbf{p}_2,\boldsymbol{\pi}_3)$ and $(\mathbf{p}_2',\boldsymbol{\pi}_3')$ are tailored to achieve optimal ride comfort and stability for each driver profile.}
    \label{fig:CaseStudyFramework}
\end{figure*}

During implementation, the DT acts as an intermediary that continuously synchronizes the digital and physical systems. Through real-time data acquisition and bidirectional communication, the digital model is updated to reflect the actual behavior of the physical system, producing an improved representation~$\mathcal{M}_1$. This learning process incorporates uncertainty quantification (UQ) and data-driven modeling to capture nonlinearities and environmental variations that were not represented in the initial model.
At the same time, the control policy is refined based on the incoming data, leading to a self-evolving feedback structure that adapts to changing conditions. Prior to advancing to the next generation, the updated digital model~$\mathcal{M}_1$ is used for the second CCD optimization (\textbf{Step 3}), in which both the physical design and control policy are re-optimized. This iterative refinement process continues through successive generations, ensuring that the system design and controller become progressively more adaptive, efficient, and resilient to uncertainty.

Beyond enabling progressive model-policy co-evolution, the DT framework also supports customization of active suspension systems for different driving behaviors. As depicted in Fig.~\ref{fig:CaseStudyFramework}, driver behavior significantly influences suspension dynamics and performance objectives. For example, a mild driver typically generates smoother road excitations and moderate acceleration profiles, whereas an aggressive driver induces sharper transients and larger dynamic loads. After the first CCD optimization, the framework diverges into separate learning pathways for each driver type. Through physical data collection, real-time model and policy updates, and subsequent redesign with the updated models and policies, the framework yields tailored solutions for each driving profile. Specifically, the DTs evolve from $\mathcal{M}_0\rightarrow\mathcal{M}_1$ for the mild driver and $\mathcal{M}_0\rightarrow\mathcal{M}_1'$ for the aggressive driver. Re-optimization using these refined models produces the design-controller pairs $(\mathbf{p}_2,\boldsymbol{\pi}_3)$ and $(\mathbf{p}_2',\boldsymbol{\pi}_3')$, which are customized to achieve optimal ride comfort, handling stability, and energy efficiency corresponding to each driving style.

Although the physical vehicle remains identical for both drivers, the resulting digital models $\mathcal{M}_1$ (mild) and $\mathcal{M}_1'$ (aggressive) differ because the quantile-based discrepancy model is updated using data collected under distinct driving behaviors, detailed in Section \ref{subsec4_4}. Each driver generates unique dynamic excitations and statistical distributions of suspension responses—smooth. For instance, a mild driver generates smoother, low-frequency inputs with limited suspension travel, whereas an aggressive driver produces large-amplitude, high-frequency excitations and stronger coupling between lift, pitch, and roll motions. Consequently, the learned discrepancy functions calibrate the DTs within their respective operational regimes to generate behaviorally specialized models. These differences reflect context-dependent digital representations of the same physical system, enabling personalized DTs that accurately capture driver-specific dynamics and improve predictive fidelity for subsequent co-design and control optimization.

This multi-generation digital-twin-enabled CCD framework therefore provides a unified methodology for continuous learning and adaptation to bridge model fidelity, real-time control, and design optimization to realize intelligent and personalized suspension systems for varying operational scenarios.

\subsection{Step 0: Initialization}
\label{subsec4_2}

Prior to training the neural networks for the control policy, an initial set of control actions is generated to provide a stable starting point for the DRL optimization. This process serves as a warm start, in which state-action pairs derived from classical control laws are used to initialize the learning-based controller~\cite{tsai2022design}. To compute approximate optimal actions for these sampled pairs under the nonlinear system dynamics, a set of proportional controllers (P controllers) is employed to regulate all state variables. The controller gains are tuned using {Bayesian Optimization} (BO) \cite{chen2024latent}, which efficiently explores the continuous gain space while balancing exploration and exploitation.

The optimization problem is formulated to minimize the overall ride discomfort, quantified by the root mean square (RMS) value of the comfort index, defined in Eq.~(\ref{eq:comfort_index}). The design vector consists of five controller parameters,
\begin{equation}
    \mathbf{k} = [K_0,\,K_1,\,K_2,\,K_3,\,K_4],
\end{equation}
which compactly represent the proportional gains for the observed states $\mathbf{y}$, including $\dot{z}_s$, $\dot{\alpha}$, $\dot{\beta}$, $z_{ui}$, and $z_s-z_{ui}$, for $i=1,...,4$. For each sampled gain vector $\mathbf{K}\in\mathbb{R}^{4\times11}$, defined as:
\begin{equation}
\resizebox{0.45\textwidth}{!}{$
\mathbf{K} =
\begin{bmatrix}
K_0 &  K_1 &  K_2 &  K_3 &  0   &  0   &  0   &  K_4 &  0   &  0   &  0 \\
K_0 & -K_1 &  K_2 &  0   &  K_3 &  0   &  0   &  0   &  K_4 &  0   &  0 \\
K_0 &  K_1 & -K_2 &  0   &  0   &  K_3 &  0   &  0   &  0   &  K_4 &  0 \\
K_0 & -K_1 & -K_2 &  0   &  0   &  0   &  K_3 &  0   &  0   &  0   &  K_4
\end{bmatrix},
$}
\label{eq:K_matrix}
\end{equation}
where each row corresponds to one actuator input ($u_1$ to $u_4$) and each column corresponds to a specific state feedback term, the system is simulated under the closed-loop law $\mathbf{u}=-\mathbf{K}\mathbf{y}$, and the corresponding comfort index is evaluated. BO then updates its surrogate model to propose the next candidate $\mathbf{k}$ that minimizes the comfort index.

After 120 iterations, the optimizer identifies the best set of gains
$\mathbf{k}^* = [5000.0,3000.0,801.3,10000.0,-1717.9]$ that minimizes the RMS comfort index, providing a physically reasonable and dynamically stable controller. This P-controller serves as the initial policy for the DRL-based CCD training, ensuring stable exploration and preventing divergence in the early learning stages.

We acknowledge that using the simplified controllers introduces approximation error and may yield suboptimal actions relative to the true nonlinear optimum. However, this approach remains valuable for high-dimensional systems. In particular, it provides a stable and informed initialization for the controller, helping to prevent instability and poor performance in the early stages of DRL training.

To initialize the DRL policy and value function, we employed fully connected feedforward neural networks with three hidden layers. The input to both networks is a 13-dimensional vector comprising 2 physical design parameters and 11 observed states. The policy network produces a 4-dimensional output corresponding to the control actions, while the value network outputs a single scalar value representing the state-value function. Each hidden layer in both networks contains 128 neurons with the hyperbolic tangent (\texttt{Tanh}) activation function applied after each layer to introduce nonlinearity and ensure smooth gradients. For the policy, two separate DNNs were used to model the mean and standard deviation of the action distribution. The samples along with the optimal actions $(\mathbf{x}_j, \mathbf{p}_j,\mathbf{u}^*_j)$ are used to initialize the mean DNN, whereas the weights and biases of the standard deviation DNN are initialized to zero and a small positive value (0.01) across all layers, respectively, to encourage initial exploration without introducing high variance.

\subsection{Step 1: First CCD Optimization}
\label{subsec4_3}
\begin{figure}[t]
    \centering
    \includegraphics[width=\linewidth]{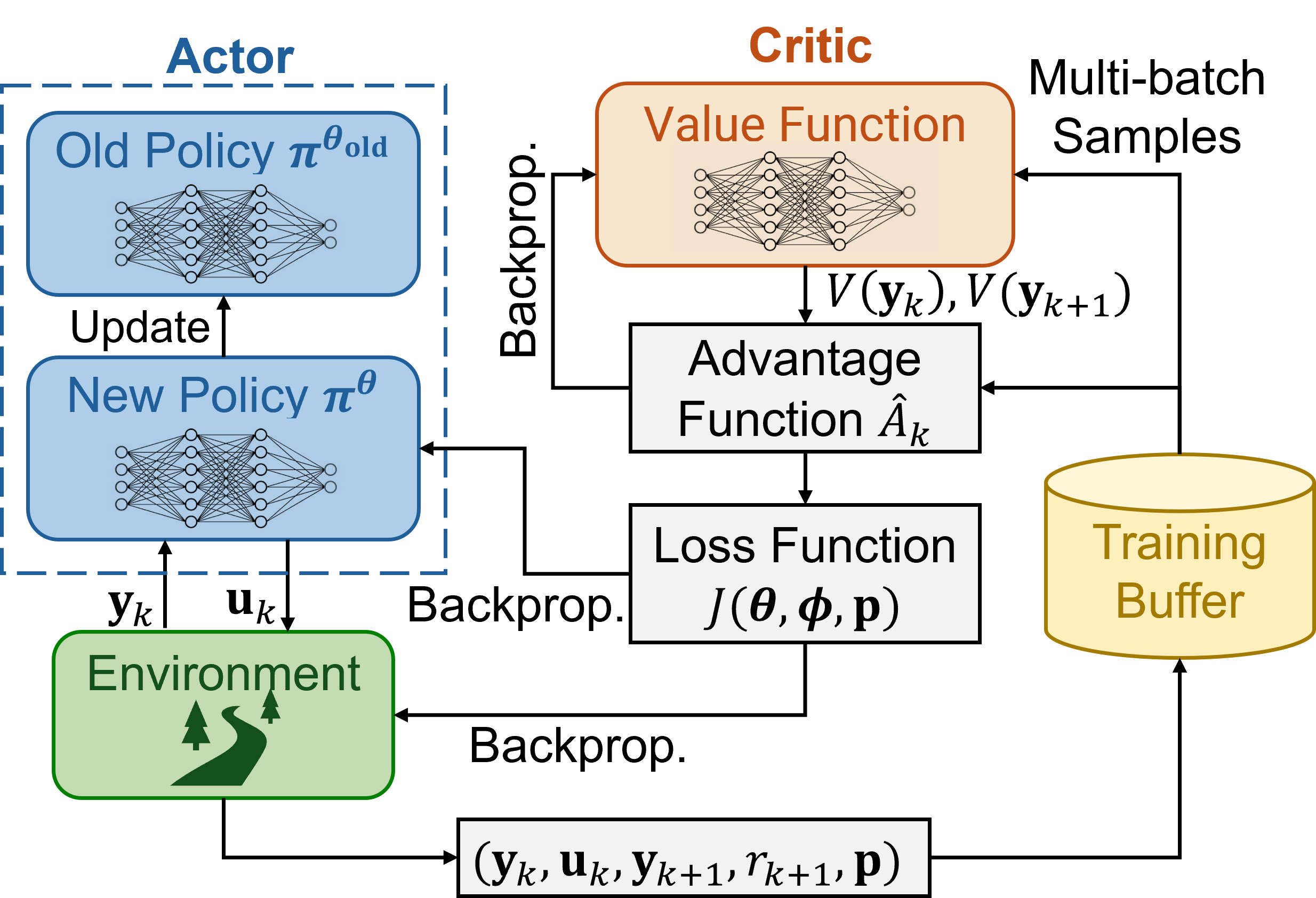}
    \caption{Flow chart of the DRL-based CCD optimization using the PPO algorithm proposed by \cite{tsai2025digital}. Unlike standard PPO implementations, the proposed framework embeds the physical design parameters $\mathbf{p}$ into the input space, allowing simultaneous gradient-based updates of both the control policy and the physical design by treating the environment dynamics as differentiable and enabling backpropagation through the environment. The policy and value networks are updated using observed states $\mathbf{y}$ since the full-vehicle suspension system is partially observable.}
    \label{fig:CCD_PPO}
\end{figure}

Following the initialization of the policy and value networks, the first stage of the multi-generation digital-twin framework performs CCD of the full-vehicle active suspension system using a DRL algorithm based on PPO. The overall structure of the PPO-based CCD optimization process is illustrated in Fig.~\ref{fig:CCD_PPO}, which shows the interaction between the actor–critic networks, the environment, and the embedded physical design parameters.

Unlike standard PPO implementations that update only the neural-network parameters of the policy and value functions, the proposed DRL-based CCD formulation jointly optimizes both the physical suspension parameters (including the stiffness coefficient $k_s$ and damping coefficient $c_s$ of all four suspension systems) and the control policy parameters. This simultaneous optimization allows the physical system and its controller to co-evolve within the same learning loop, ensuring that the control policy adapts to hardware changes while the design variables are refined to improve closed-loop performance.

The CCD problem minimizes a joint loss function composed of a clipped surrogate policy objective and the value loss:

\begin{align}
   \min_{\boldsymbol\theta, \boldsymbol\phi, \mathbf{p}}\; &J(\boldsymbol\theta, \boldsymbol\phi, \mathbf{p})\notag\\
   =\mathbb{E}_k &\left[
\underbrace{-\min \left( \rho_k(\boldsymbol\theta, \mathbf{p}) \hat{A}_k,L^ \text{CLIP}(\rho_k(\boldsymbol\theta, \mathbf{p}), 1-\epsilon, 1+\epsilon)\hat{A}_k \right)}_{\text{Policy Loss}}\right.\notag\\
&\left.+ c_v \cdot \underbrace{L_{\text{SmoothL1}}\left(V^{\boldsymbol\phi}(\mathbf{x}_k, \mathbf{p}),\hat{V}_k\right)}_{\text{Value Loss}}\right],
\label{eq:PPO_loss}
\end{align}
where $\hat{A}_k$ is the advantage function at time step $k$, which estimates how much better an action is compared to the expected value of a state, and $L^ \text{CLIP}(\rho_k(\boldsymbol{\theta},\mathbf{p}),1-\epsilon,1+\epsilon)$ is a function clipping $\rho_k(\boldsymbol{\theta},\mathbf{p})$ within $(1-\epsilon,1+\epsilon)$, $\rho_k$ denotes the probability ratio between the new and old policies:
\begin{equation}
    \rho_k(\boldsymbol{\theta},\mathbf{p})=\frac{\boldsymbol{\pi}^{\boldsymbol\theta}\left(\mathbf{u}_k|\mathbf{x}_k,\mathbf{p}\right)}{\boldsymbol{\pi}^{\boldsymbol\theta_{\text{old}}}\left(\mathbf{u}_k|\mathbf{x}_k,\mathbf{p}_{\text{old}}\right)},
\end{equation}
$c_v$ is the coefficient for value loss, and we use Smooth L1 loss to define the value loss:
\begin{equation}
    L_{\text{SmoothL1}}(a, b) = 
    \begin{cases}
        \frac{1}{2}(a - b)^2, & \text{if } |x - y| < 1 \\
        |a - b| - \frac{1}{2}, & \text{otherwise.}
    \end{cases}
\end{equation}

By embedding the suspension design vector $\mathbf{p}$ (e.g., $k_s$ and $c_s$ for each suspension) into both policy and value networks, the optimization becomes differentiable with respect to the physical parameters, along with the parameters of the neural networks for the policy and value function. Gradients are computed through automatic differentiation using \texttt{PyTorch} autograd engine~\cite{paszke2019pytorch}, enabling end-to-end updates of all trainable variables. This mechanism effectively treats the entire digital-twin model, including vehicle dynamics, controller, and design parameters, as a single computational graph, allowing the optimizer to explore both mechanical-design and control-policy spaces concurrently.

During training, the algorithm seeks to maximize ride comfort and handling stability by minimizing body acceleration, pitch/roll motion, and control effort. During training, the full-vehicle model is simulated under a constant forward velocity of 10 m/s, zero acceleration, and zero steering angle, representing a straight-line driving scenario. To emulate random road irregularities, the road disturbance $z_{r,i}$ and its time derivative $\dot{z}_{r,i}$ are sampled at each time step from zero-mean Gaussian distributions,
\begin{equation}
    z_{r,i}\sim\mathcal{N}(0,0.001^2),~\dot{z}_{r,i}\sim\mathcal{N}(0,0.1^2),
\end{equation}
applied independently to each of the four wheels ($i=1,2,3,$ and $4$). The reward function penalizes body acceleration, pitch and roll rates, and control effort, driving the system toward an optimal trade-off between ride comfort, handling stability, and actuator energy efficiency.

\begin{figure}
    \centering
	\begin{subfigure}{0.49\textwidth}
        \centering
    	\includegraphics[width=\textwidth]{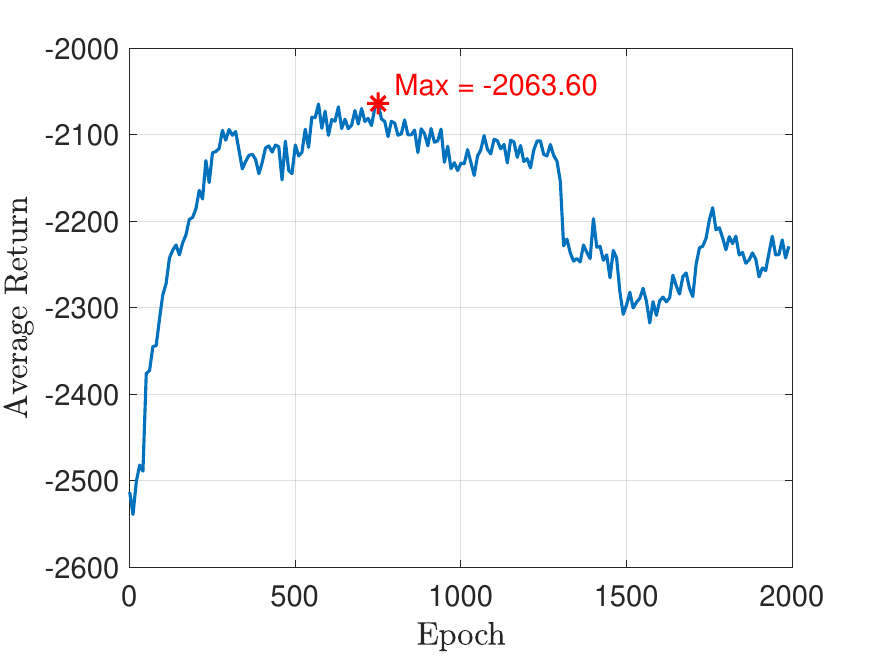}
    	\caption{}
    	\label{fig:history_return_first_opt}
	\end{subfigure}
    \hfill
	\begin{subfigure}{0.49\textwidth}
        \centering
    	\includegraphics[width=\textwidth]{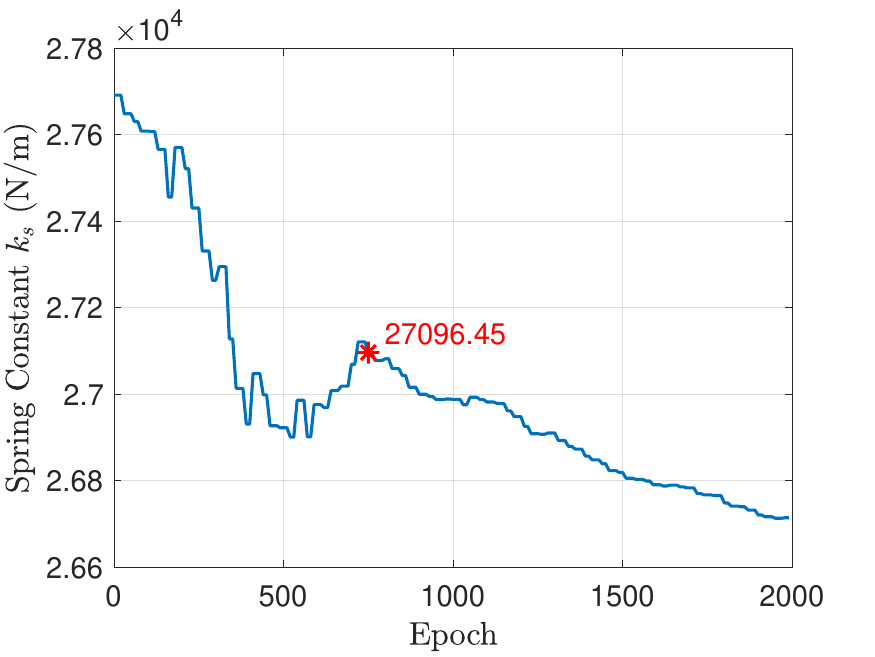}
    	\caption{}
    	\label{fig:history_ks_first_opt}
	\end{subfigure}
    \hfill
	\begin{subfigure}{0.49\textwidth}
        \centering
    	\includegraphics[width=\textwidth]{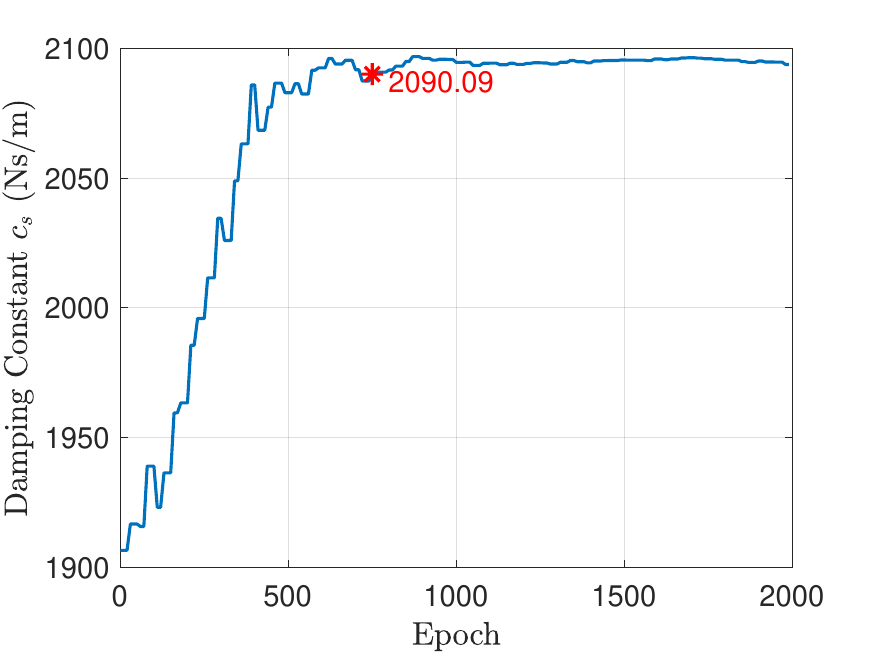}
    	\caption{}
    	\label{fig:history_cs_first_opt}
	\end{subfigure}
    \caption{Training history for the first CCD optimization with (a) average return, (b) system parameter: spring constant $k_s$, and (c) system parameter: damping constant $c_s$.}
    \label{fig:history_first_opt}
\end{figure}

Figure \ref{fig:history_first_opt} illustrates the training history, where the average return increases from approximately –2600 to –2063.6 over 2000 epochs (Fig. \ref{fig:history_return_first_opt}). The spring stiffness and damping coefficients converge to $k_s=27096.45$ N/m and $c_s=2090.09$ N$\cdot$s/m, shown in Fig. \ref{fig:history_ks_first_opt} and Fig. \ref{fig:history_cs_first_opt}, respectively. These results demonstrate how the DRL-based CCD framework successfully balances stiffness and damping to suppress body motion while maintaining responsive control actions. The optimized design vector $\mathbf{p}_1$ and $\boldsymbol{\pi}_1$ define the Generation-1 configuration, which will be deployed in the next step for physical implementation and data-driven model updating.

Although this simulation setup may appear simplified and not fully generalizable to diverse real-world scenarios, it represents a necessary first step before physical implementation, as it requires no experimental data or real-road measurements. The Gaussian road disturbances introduce controlled stochasticity and uncertainty, allowing the learning agent to experience varying excitation patterns without relying on physical tests.

Compared with many existing studies (e.g., \cite{dridi2025optimizing}), which used only deterministic bump or rectangular-road excitations, our setup provides a more statistically representative yet computationally tractable environment for early-stage training. This design enables the framework to capture fundamental co-design behaviors and evaluate the learning stability of the DRL-based CCD algorithm before moving to the next stage, where real-world road profiles and driving behaviors are incorporated through physical implementation in the digital-twin loop.

\subsection{Step 2: First Implementation (Generation 1)}
\label{subsec4_4}

Before deploying the optimized design and policy from Step 1 to the real environment, it is essential to recognize and quantify the mismatch between the digital model and the physical vehicle system. Even with high-fidelity modeling, discrepancies inevitably arise from nonlinearities not being captured, component variability, and environmental factors not represented in the nominal digital model. In practice, these mismatches lead to performance degradation if the control policy is transferred directly without adaptation.

To emulate such discrepancies, we construct a hypothetical real system that introduces realistic deviations from the nominal simulation model in three primary ways:

\begin{enumerate}
    \item \textbf{Nonlinear spring and damper behavior.}
    In real suspensions, stiffness and damping characteristics are rarely linear across the full stroke and velocity range. 
    To account for this, the spring and damper forces are augmented with cubic and velocity-dependent nonlinear terms:
    \begin{align}
        F_{S_i} = &k_s(\,z_{u,i}-z_s-\Delta_i\,) 
        + c_s(\,\dot{z}_{u,i}-\dot{z}_s-\dot{\Delta}_i\,)
        \notag\\+ &k_{\text{nl}}(\,z_{u,i}-z_s-\Delta_i\,)^3\notag
        \\+ &c_{\text{nl}}\big|\dot{z}_{u,i}-\dot{z}_s-\dot{\Delta}_i\big|
        (\dot{z}_{u,i}-\dot{z}_s-\dot{\Delta}_i),
        \label{eq:nonlinear_force}
    \end{align}
    where $k_{\text{nl}} = 0.1k_s$ and $c_{\text{nl}} = 0.1c_s$. 
    This nonlinear formulation captures the amplitude-dependent stiffness and rate-sensitive damping observed in real suspension systems.

    \item \textbf{Non-uniform unsprung masses.}
    Manufacturing variability, tire-wheel assemblies, and sensor packages introduce asymmetric wheel masses. 
    The unsprung mass vector is therefore defined as
    \begin{equation}
        \mathbf{m}_u = [60.0,\, 50.0,\, 45.0,\, 50.0]~\text{kg},
    \end{equation}
    representing the front-left, front-right, rear-left, and rear-right wheels, respectively.

    \item \textbf{Asymmetric vehicle geometry.}
    To reflect potential variations in the structural layout and load distribution, the center-of-gravity height is perturbed by $+0.05~\text{m}$ relative to the nominal model, giving
    \begin{equation}
        h_{\text{cg}} = 0.55 + 0.05~\text{m}.
    \end{equation}
    This asymmetry slightly shifts the pitch and roll dynamics, leading to differential responses across the front and rear suspensions.
    
    \item \textbf{Heterogeneous tire stiffnesses across wheels.}
    Real tires exhibit corner-dependent stiffness due to wear, load, and construction differences. 
    We model this by assigning wheel-specific tire stiffnesses
    \begin{equation}
        \mathbf{k}_t = [0.9,\,1.2,\,1.1,\,0.9]\times 200{,}000~\text{N/m},
    \end{equation}
    for the front-left, front-right, rear-left, and rear-right wheels, respectively. 
    The tire forces are then
    \begin{align}
        F_{T_i} = k_{t,i}\big(z_{r,i}-z_{u,i}\big) + c_t\big(\dot{z}_{r,i}-\dot{z}_{u,i}\big).
        \label{eq:tire_forces}
    \end{align}

    \item \textbf{Deviations in sprung mass and moments of inertia.}
    To represent loading variation and structural uncertainty, the sprung mass and rotational inertias are increased by 10\% relative to the nominal values:
    \begin{align}
        m_s = 1.1\times 1500 = 1650~\text{kg},\\
        I_{\theta} = 1.1\times 2500 = 2750~\text{kg}\cdot\text{m}^2,\\
        I_{\phi} = 1.1\times 500 = 550~\text{kg}\cdot\text{m}^2.
    \end{align}
    These deviations alter the heave--pitch--roll couplings and shift natural frequencies, increasing the reality gap for controller transfer.
\end{enumerate}
Together, these modifications define the {Generation~1 physical system}, which serves as the ``real'' environment in the digital-twin framework. The policy $\boldsymbol\pi_1$ and design parameters $\mathbf{p}_1$ obtained from \textbf{Step 1} are directly deployed in this perturbed environment to evaluate {transfer performance} under realistic uncertainty. 


\begin{figure*}
    \centering
	\begin{subfigure}{0.49\textwidth}
        \centering
    	\includegraphics[width=\textwidth]{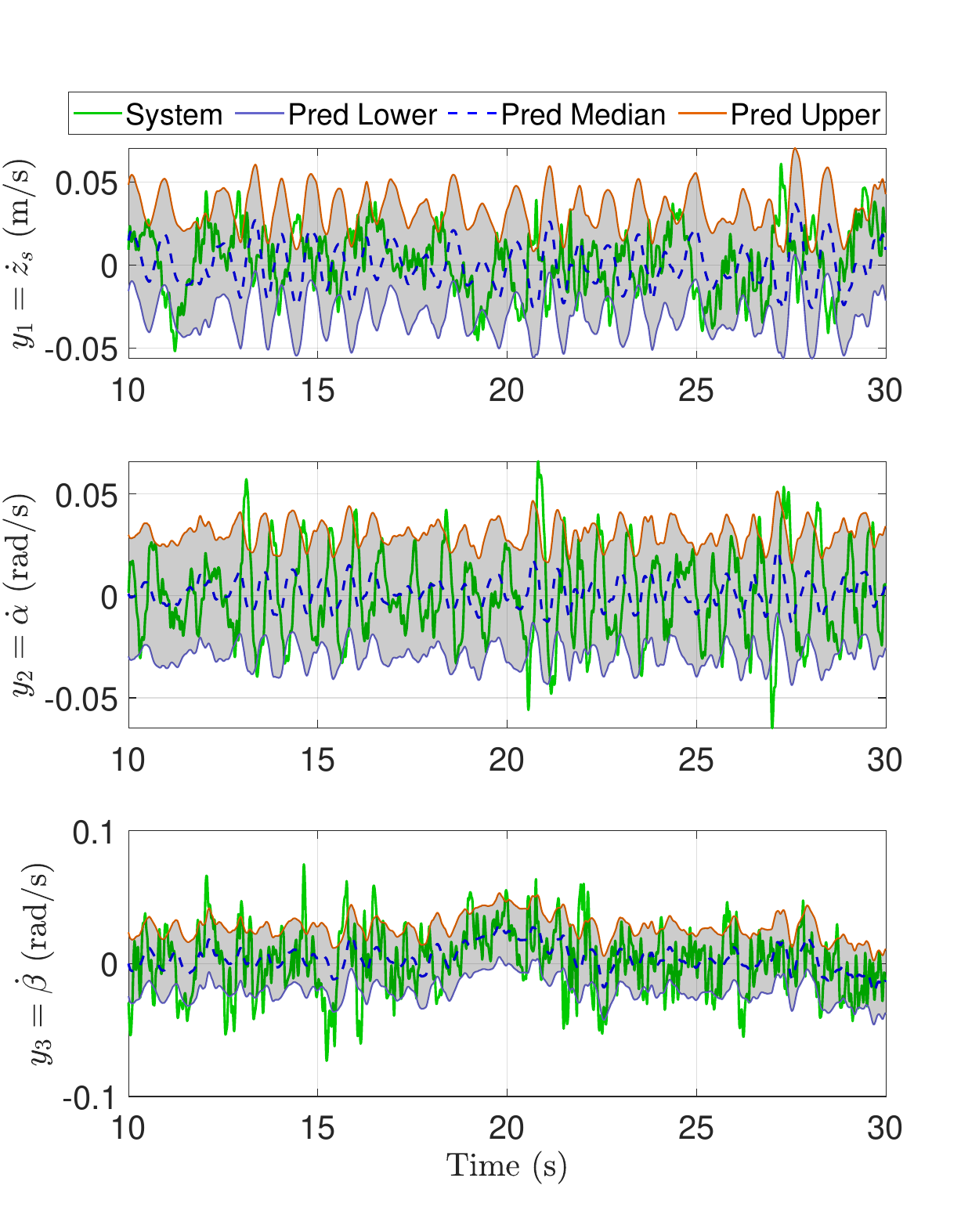}
    	\caption{Mild driver.}
    	\label{fig:quantile_mild}
	\end{subfigure}
    \hfill
	\begin{subfigure}{0.49\textwidth}
        \centering
    	\includegraphics[width=\textwidth]{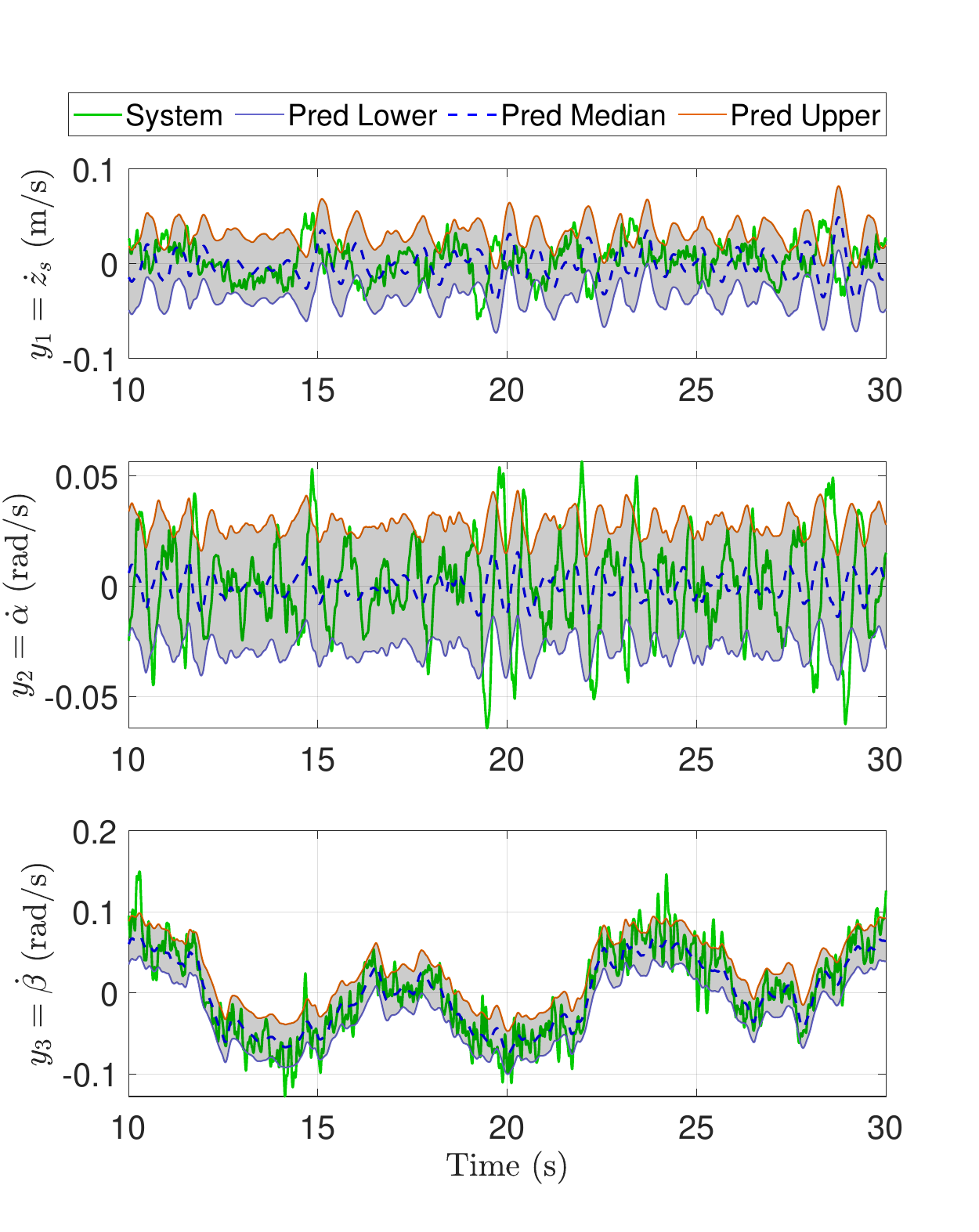}
    	\caption{Aggressive driver.}
    	\label{fig:quantile_aggr}
	\end{subfigure}
    \caption{Visualization of the learned quantiles and the
real system trajectories of the first three observed states for the active suspension system.}
    \label{fig:quantiles}
\end{figure*}

Despite the discrepancies between the digital model and the physical system, the DRL-based controller demonstrates strong robustness and adaptability. This resilience arises from the inherently adaptive nature of RL, which enables the policy to evolve in response to changes in system dynamics and environmental conditions. Within the digital-twin framework, sensor data collected from the physical vehicle (e.g., suspension deflections, actuator forces, and body accelerations) are continuously streamed to the digital model. These real-time observations are used to update the policy ($\boldsymbol{\pi}_2$). This allows the agent to refine its decision-making strategy based on the latest system behavior. Through this process, the RL agent learns to maintain closed-loop stability and optimal ride–handling balance even under uncertainty by gradually compensating for model–reality discrepancies, nonlinear effects, and asymmetric parameter variations introduced in the physical system.

To explicitly account for the mismatch between the digital and physical systems, a discrepancy model is developed using physical data collected during the Generation 1 deployment. This model quantifies deviations between predicted and observed system responses by leveraging quantile regression \cite{chen2025uncertainty}, a data-driven uncertainty quantification (UQ) technique that estimates conditional quantiles of a response variable. Unlike traditional regression methods, quantile regression provides predictive intervals that mainly capture the uncertainties from biased parameters and environmental noises~\cite{chen2025uncertainty}. Moreover, this approach does not need to assume Gaussian noise, thereby improving its flexibility in representing nonlinear, asymmetric errors commonly observed in real vehicle dynamics. Different from online calibrating model parameters \cite{chen2021unknown}, we argue that a discrepancy model directly applied to correct the system dynamic model allows better generalization while capturing missing physics. 

In practice, the DT continuously collects real-time data from onboard sensors to monitor the vehicle state as well as driving conditions such as forward acceleration, vehicle speed, and steering angle. These measurements provide the essential information required to identify and update the discrepancy between the digital model and the real system. Through learning from this data stream, the discrepancy model dynamically refines its estimates of modeling error during operation.

By training the model at representative quantile levels, typically the 10th, 50th (median), and 90th percentiles, the DT captures both the central tendency and spread of the discrepancy distribution. The discrepancy at the next time step is estimated as a function of the current observed states, control actions, previous errors, and the driving conditions (forward acceleration $a$, speed $v$, and steering angle $\delta$):

\begin{equation}
    \begin{bmatrix}\mathbf{e}^{\text{upper}}_{k+1}\\\mathbf{e}^{\text{median}}_{k+1}\\\mathbf{e}^{\text{lower}}_{k+1}\end{bmatrix}=\mathbf{f_e}(\mathbf{e}_k,\mathbf{y}_k,\mathbf{u}_k,a_{k},v_{k},\delta_{k}),
    \label{eq:quantile_equation}
\end{equation}
where $\mathbf{e}_k := \mathbf{y}_k - \hat{\mathbf{y}}_k$ denotes the deviation between the actual state $\mathbf{y}_k$ and the nominal predicted state $\hat{\mathbf{y}}_k$ from the digital model. The predicted quantiles $\mathbf{e}^{\text{upper}}_{k+1}$, $\mathbf{e}^{\text{median}}_{k+1}$, and $\mathbf{e}^{\text{lower}}_{k+1}$ represent the range of the deviations at the next step.

As shown in Fig.~\ref{fig:quantiles}, the learned quantiles effectively bound the observed trajectories of the first three system states and capture both the variability and the bias between the digital model predictions and the real physical responses. The root mean square errors for the validation dataset are 0.0147 and 0.0151 under mild and aggressive driving profiles, respectively. The median quantile tracks the mean trajectory, while the upper and lower quantiles define the confidence region representing the dynamic uncertainty envelope of the physical system. This learned discrepancy model is then integrated into the DT to adaptively calibrate predictions and guide the controller update for subsequent generations.

While the learned quantiles successfully capture most of the real trajectories, a few observed data points fall outside the predicted uncertainty bounds, particularly in the third state corresponding to the roll velocity $\dot{\beta}$ (see Fig.~\ref{fig:quantiles}). This behavior stems from an unbalanced dataset, where the frequency of extreme operating conditions is considerably lower than that of regular driving states. As a result, the quantile regression model tends to underestimate the probability of rare events and overestimate uncertainty during nominal conditions. Nevertheless, as more diverse data become available through subsequent generations of physical operation, the quantile estimates are expected to become more accurate and better calibrated across the full operational spectrum.

\begin{figure*}
    \centering
	\begin{subfigure}{0.9\textwidth}
        \centering
    	\includegraphics[width=\textwidth]{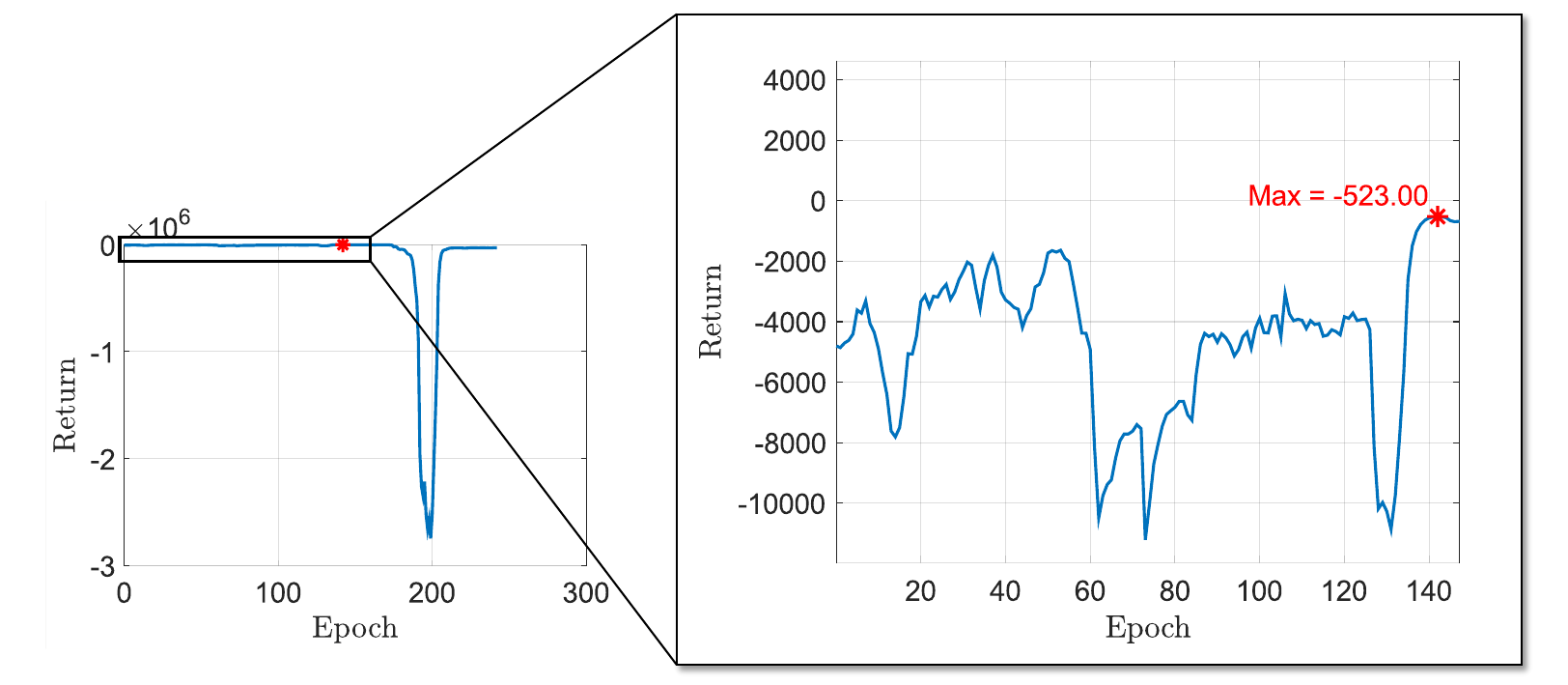}
    	\caption{}
    	\label{fig:history_return_second_opt_mild}
	\end{subfigure}
    \hfill
	\begin{subfigure}{0.49\textwidth}
        \centering
    	\includegraphics[width=\textwidth]{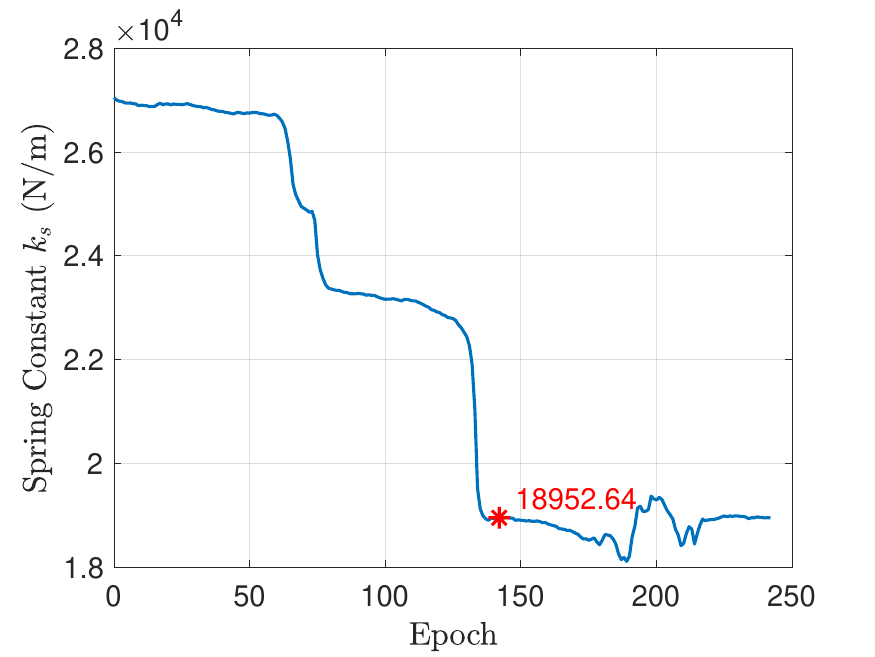}
    	\caption{}
    	\label{fig:history_ks_second_opt_mild}
	\end{subfigure}
    \hfill
	\begin{subfigure}{0.49\textwidth}
        \centering
    	\includegraphics[width=\textwidth]{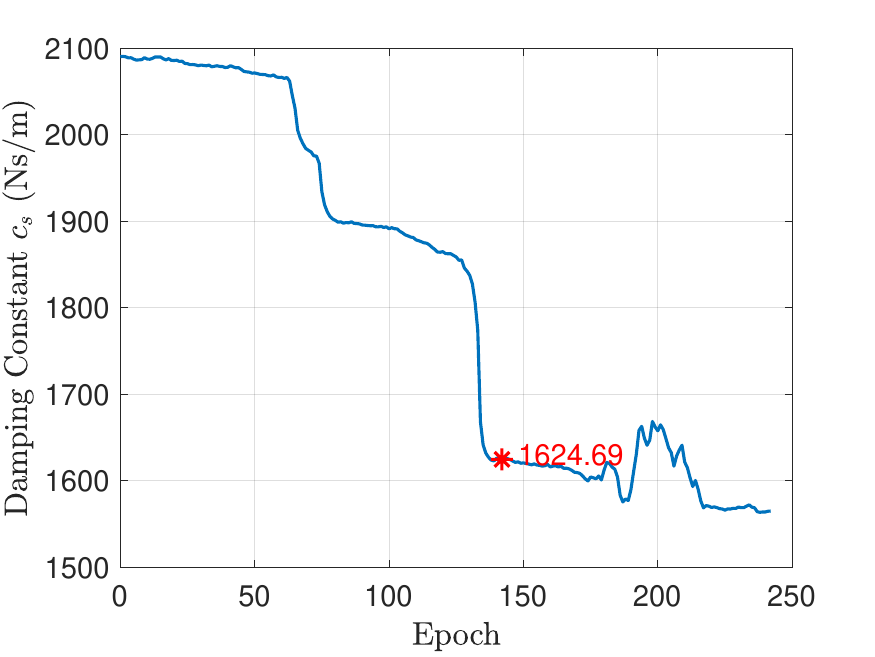}
    	\caption{}
    	\label{fig:history_cs_second_opt_mild}
	\end{subfigure}
    \caption{Training history of the second CCD optimization for the mild driver, with (a) return, (b) system parameter: spring constant $k_s$, and (c) system parameter: damping constant $c_s$.}
    \label{fig:history_second_opt_mild}
\end{figure*}

\subsection{Step 3: Second CCD Optimization}
\label{subsec4_5}

The primary distinction between the first and second CCD optimizations lies in the integration of the updated digital-twin model that incorporates the quantile-based discrepancy learning derived from Eq.~(\ref{eq:quantile_equation}). With the physical data collected during the first implementation, the digital model now provides a more accurate and uncertainty-aware representation of the vehicle dynamics. Unlike the initial optimization from \textbf{Step 1}, which relies on stochastically generated Gaussian disturbances to emulate environmental variability, the second CCD optimization directly leverages the learned discrepancy model to reflect realistic uncertainty in both the dynamics and external excitations. The updated model is constructed by incorporating the learned discrepancy function from Eq.~(\ref{eq:quantile_equation}) into Eq.~(\ref{eq:discrete_dynamics}), defined as:
\begin{align}
\begin{bmatrix}\mathbf{y}^{\text{upper}}_{k+1}\\ \mathbf{y}^{\text{median}}_{k+1}\\\mathbf{y}^{\text{lower}}_{k+1}\end{bmatrix}
    = \mathbf{C}\mathbf{x}_{k+1}\notag\\+\mathbf{f_e}(\mathbf{e}_k,\mathbf{y}_k,\mathbf{u}_k,a_{k},v_{k},\delta_{k})
\end{align}
where $\mathbf{x}_{k+1}$ is the next state estimated by the model from Eq.~(\ref{eq:discrete_dynamics}). The optimization process no longer depends on artificially injected noise; instead, it inherently embeds the variability observed in the real system through the model. This advancement enables a more faithful simulation environment for DRL, leading to improved transferability and faster policy convergence in subsequent generations.

In the second CCD optimization, the new reward function is extended from Eq.~(\ref{eq:reward_susp}) to explicitly account for the uncertainties captured by the quantile-informed digital model, leading to:
\begin{equation}
    r_{k+1}'
    = -\Big(
        r_{k+1}
        + \lambda_u\,J_{\text{uncertainty}}
      \Big),
    \label{eq:reward_second}
\end{equation}
where
\begin{equation}
    J_{\text{uncertainty}}
    = J_{\text{comfort,unc}}
      + c_1\,J_{\text{pitch,unc}}
      + c_2\,J_{\text{roll,unc}}.
    \label{eq:reward_uncertainty}
\end{equation}
The corresponding uncertainty term is computed as
\begin{equation}
    J_{\text{comfort,unc}}
    = \sqrt{(w_1\,\Delta a_{z,k})^2 + (w_2\,\Delta \ddot{\theta}_k)^2 + (w_3\,\Delta \ddot{\phi}_k)^2},
\end{equation}
where $\Delta(\cdot)$ represents the prediction uncertainty estimated by the quantile regression model. Similarly, $J_{\text{pitch,unc}}=(\Delta\alpha_k)^2$ and $J_{\text{roll,unc}}=(\Delta\beta_k)^2$ represent their corresponding uncertainty penalties. This modified formulation encourages the agent to not only optimize nominal performance but also achieve robust behavior by minimizing the propagation of model and environmental uncertainties.

\begin{figure}
    \centering
	\begin{subfigure}{0.49\textwidth}
        \centering
    	\includegraphics[width=\textwidth]{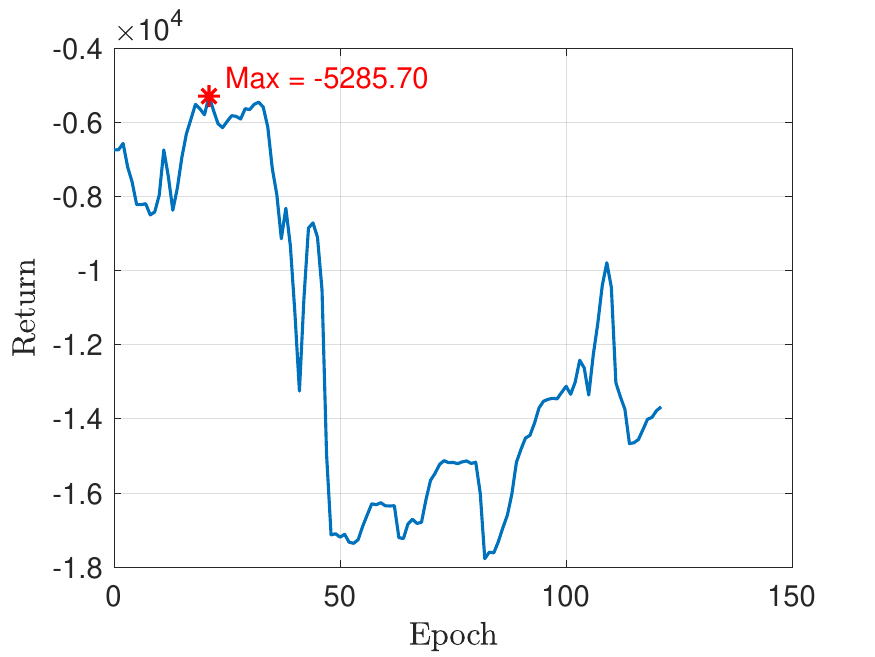}
    	\caption{}
    	\label{fig:history_return_second_opt_aggr}
	\end{subfigure}
    \hfill
	\begin{subfigure}{0.49\textwidth}
        \centering
    	\includegraphics[width=\textwidth]{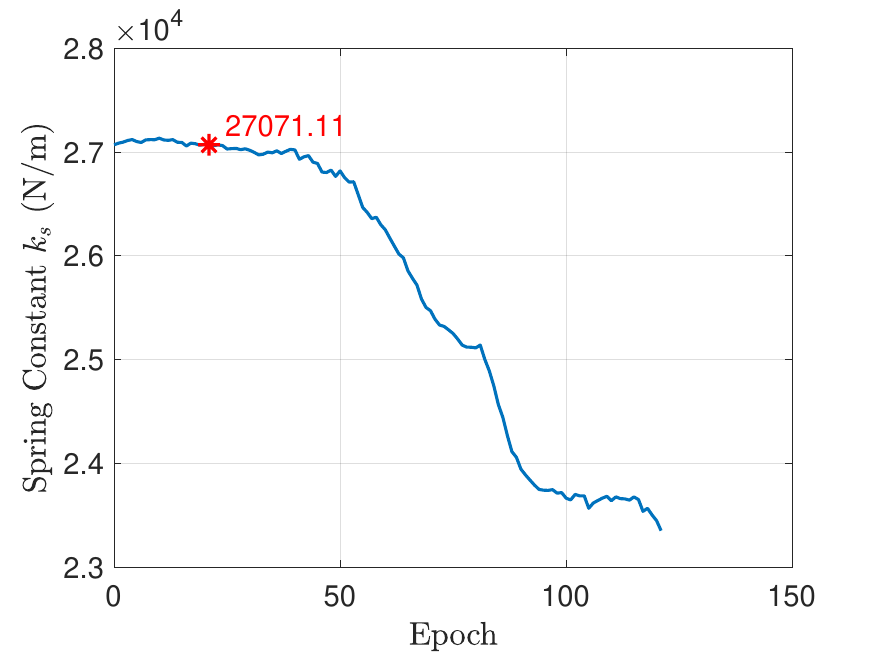}
    	\caption{}
    	\label{fig:history_ks_second_opt_aggr}
	\end{subfigure}
    \hfill
	\begin{subfigure}{0.49\textwidth}
        \centering
    	\includegraphics[width=\textwidth]{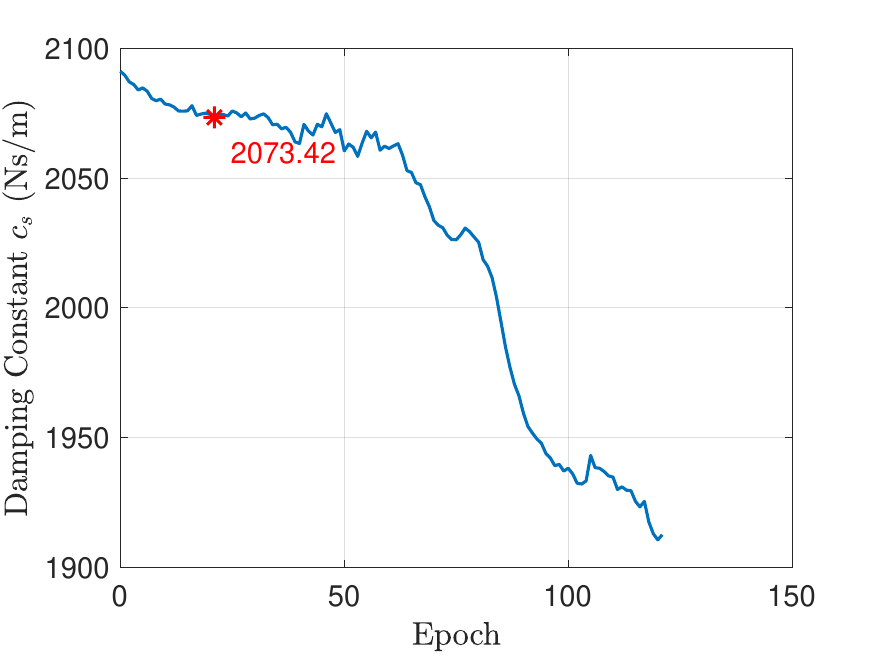}
    	\caption{}
    	\label{fig:history_cs_second_opt_aggr}
	\end{subfigure}
    \caption{Training history of the second CCD optimization for the aggressive driver, with (a) return, (b) system parameter: spring constant $k_s$, and (c) system parameter: damping constant $c_s$.}
    \label{fig:history_second_opt_aggr}
\end{figure}

Figure~\ref{fig:history_second_opt_mild} shows the training histories of the return, spring stiffness $k_s$, and damping coefficient $c_s$ for the mild driver over approximately 250 epochs. A new best model was identified at epoch~142 with a reward of $-523.00$, after which no further improvement was observed and the patience-based early stopping criterion ($\text{patience}=100$) terminated training near epoch~242. 

It is observed that the return temporarily drops to extremely low values around epochs~190--210 before recovering to its typical range (Fig.~\ref{fig:history_return_second_opt_mild}). This behavior is not uncommon in DRL-based co-design problems and can be attributed to the intrinsic stochasticity and nonstationarity of the training process. During training, the PPO agent explores the design and control spaces by sampling actions from a stochastic policy, which can occasionally generate trajectories associated with unusually poor rewards. Because each policy update is based on newly collected rollouts, such exploratory episodes may yield abrupt performance drops even when the overall learning trend remains positive. Moreover, unlike conventional control-only training, the physical design variables $(k_s,c_s)$ are optimized jointly with the control policy, introducing additional nonstationarity in the environment. When the co-design updates temporarily shift the system into a less favorable region of the design space, such as an overly soft or stiff suspension configuration, the reward may decrease sharply until the policy readapts to the new dynamics. This explains the transient return collapse observed around epoch~200, which coincides with small step changes in the design parameters. The clipped surrogate objective of the PPO algorithm and the conservative learning rate prevent these disturbances from causing divergence, allowing the training to recover and converge to a stable solution. Overall, the training demonstrates strong resilience, ultimately converging with stable design parameters and policy performance once the reward plateaued beyond epoch~142.

Figure~\ref{fig:history_second_opt_aggr} presents the training histories for the aggressive-driving scenario. Compared with the mild-driver case, the second CCD optimization converges much earlier, with the best model identified at epoch~21 and an average reward of $-5285.72$. Afterward, no further improvement was observed, and the training quickly reached the patience threshold for early stopping. The faster convergence arises because the updated digital model $\mathcal{M}_1'$ effectively captured the more energetic driving conditions. Another contributing factor is that, under aggressive driving, the system operates closer to its physical and control limits, leaving less room for further improvement through policy or design adjustments. As a result, the learning process quickly stabilizes around a near-optimal solution that balances ride comfort, stability, and control effort within the achievable performance envelope of the suspension system.

The optimized design parameters converged to $k_s = 27{,}071.11~\text{N/m}$ and $c_s = 2{,}073.42~\text{N$\cdot$s/m}$, both higher than those of the mild-driver design, which aligns with the physical intuition that stiffer and more strongly damped suspensions are required to maintain stability under aggressive driving maneuvers. These results further confirm that the multi-generation digital-twin framework can efficiently adapt and customize the suspension design and control policy for distinct driving behaviors.

\begin{table*} 
\centering 
\caption{Summary of initial and optimized suspension design parameters ($k_s$: spring constant and $c_s$: damping constant).} 
    \begin{tabular}{lll} 
    \hline  & {$k_s$} (N/m) & {$c_s$} (N$\cdot$s/m) \\ \hline 
    Initial & 27692.00 & 1906.50 \\
    Optimized after first CCD & 27096.45 & 2090.09 \\
    Optimized after second CCD (mild) & 18952.64 & 1624.69\\
    Optimized after second CCD (aggressive) & 27071.11 & 2073.42\\ \hline 
    \end{tabular} 
\label{tab:initial_optimized_designs} 
\end{table*}


\begin{table*} 
\centering 
\caption{Performance comparison before and after the {second} CCD optimization for mild and aggressive drivers. Improvements are computed as $|\text{Before}-\text{After}|/\text{Before}\times 100\%$.} 
    \begin{tabular}{lllll} 
    \hline & {Metric} & {Before} & {After} & {Improvement} \\ \hline 
    \multirow{2}{*}{Mild} & RMS of body acceleration (${\text{m/s}^2}$) & 0.271812 & 0.249357 & 8.26\% \\ & Control effort (Mean of $|\mathbf{u}_k|$, N) & 260.167 & 147.541 & \textbf{43.29\%} \\ \hline 
    \multirow{2}{*}{Aggressive} & RMS of body acceleration (${\text{m/s}^2}$) & 0.267725 & 0.275832 & - \\ & Control effort (Mean of $|\mathbf{u}_k|$, N) & 315.962 & 151.474 & \textbf{52.06\%} \\ \hline 
    \end{tabular} 
\label{tab:returns_step3_susp} 
\end{table*}

\begin{figure*}[t]
    \centering
    \includegraphics[width=\linewidth]{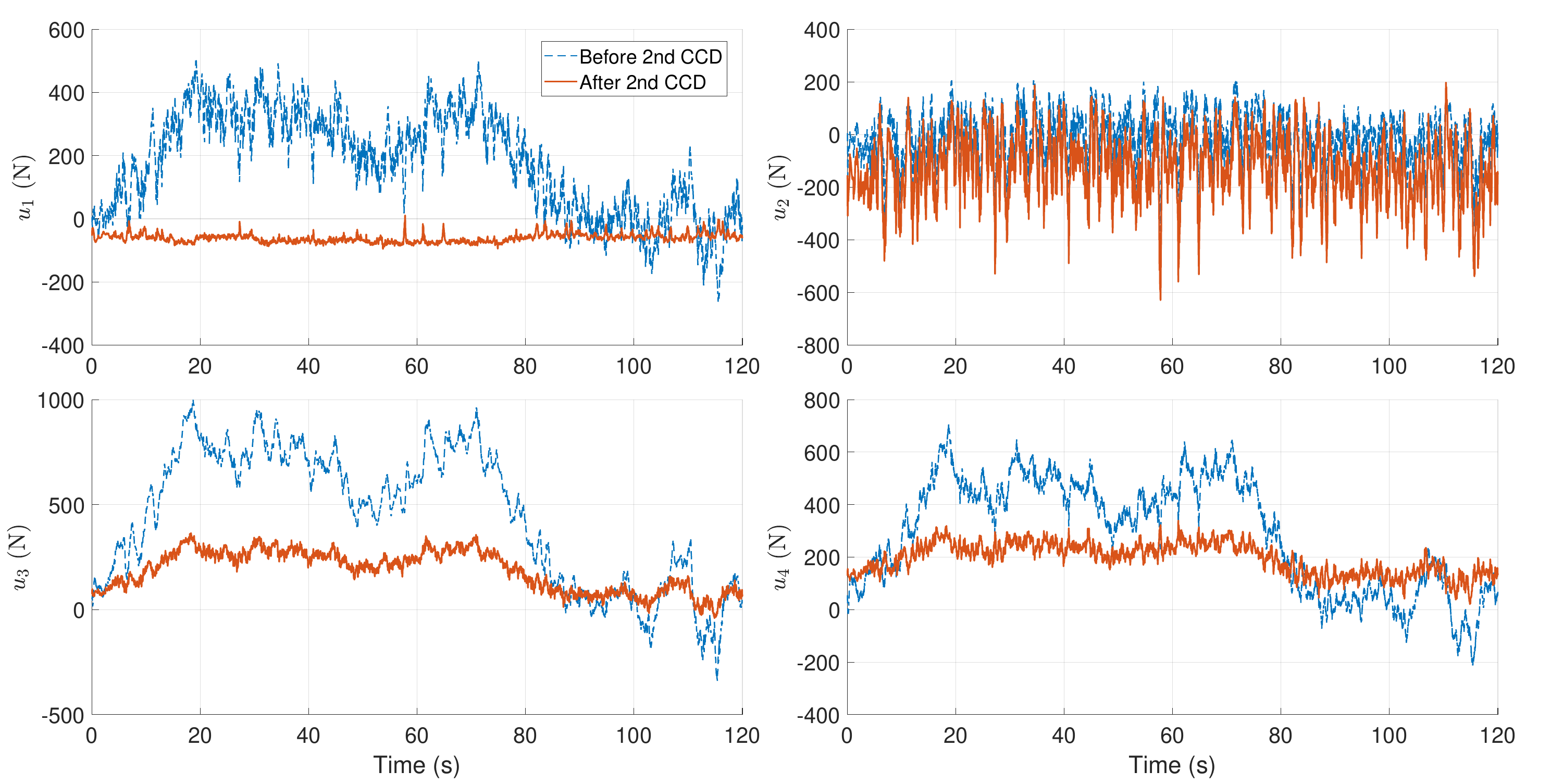}
    \caption{Control trajectories of the mild-driver scenario before and after the second CCD optimization. Only the first 120 seconds of the trajectories are shown for clarity.}
    \label{fig:action_traj_mild}
\end{figure*}

\begin{figure*}[t]
    \centering
    \includegraphics[width=\linewidth]{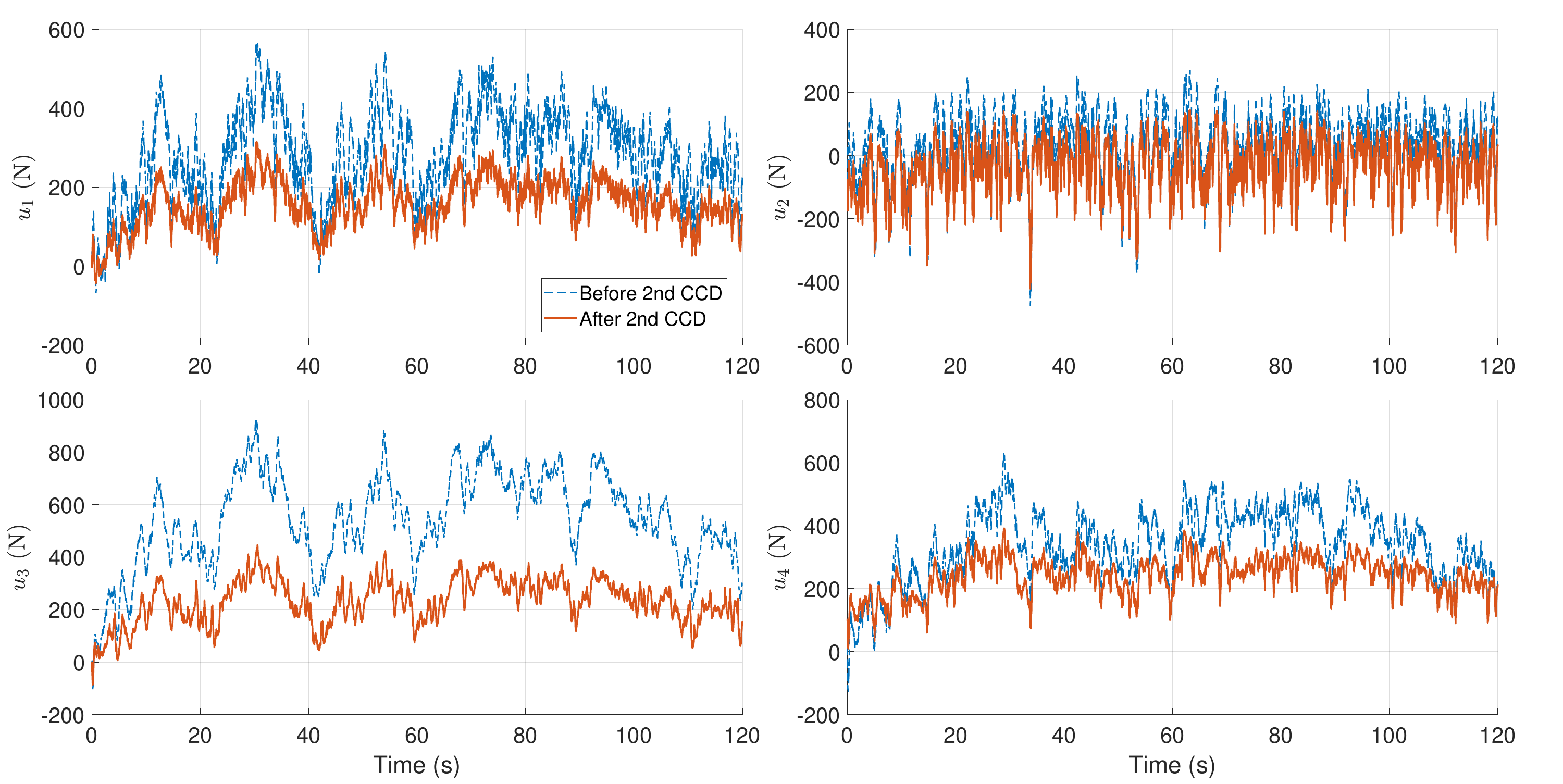}
    \caption{Control trajectories of the aggressive-driver scenario before and after the second CCD optimization. Only the first 120 seconds of the trajectories are shown for clarity.}
    \label{fig:action_traj_aggr}
\end{figure*}

\section{Results and Discussion}\label{sec5}

\subsection{Suspension Designs}
Table~\ref{tab:initial_optimized_designs} summarizes the spring and damping constants (\(k_s, c_s\)) for different stages of the CCD process (initial, optimized after first CCD, and optimized after second CCD for two driving scenarios). These parameters collectively define the passive dynamics of the suspension, which interact with the active control system to achieve desired ride and handling performance. The relationship between the physical suspension design and the control system can be interpreted through the framework of open-loop and closed-loop co-design \cite{deshmukh2015bridging}. A softer suspension (lower \(k_s\) and \(c_s\)) implies that the passive system alone provides less vibration suppression and stability; however, this allows the active controllers to play a larger role in compensating external disturbances and maintaining ride comfort. Conversely, a stiffer suspension limits the range of active control action but enhances the inherent stability and responsiveness of the vehicle.

For the mild-driver scenario, the second CCD optimization results in significantly reduced stiffness and damping (\(k_s = 18952.64~\mathrm{N/m}\), \(c_s = 1624.69~\mathrm{N\cdot s/m}\)). This softer configuration indicates that the control system effectively assumes a greater share of the vibration mitigation task, actively stabilizing the vehicle and isolating the body motion from road irregularities. Such a design is physically intuitive, as mild driving introduces smaller disturbances, allowing the actuators to manage vibrations without large control effort while maintaining comfort.

In contrast, the aggressive-driving scenario exhibits a much stiffer configuration (\(k_s = 27071.11~\mathrm{N/m}\), \(c_s = 2073.42~\mathrm{N\cdot s/m}\)), reflecting the higher demand for dynamic stability and fast response under harsh maneuvers. The system favors a more rigid suspension that provides stronger passive support to resist large load transfers and rapid body oscillations, thereby reducing the burden on the controller. This trend aligns with the intuition from the active-passive co-design literature~\cite{deshmukh2015bridging}, where aggressive operating conditions drive the optimization toward increased structural stiffness and damping to maintain controllability and robustness. Overall, these results reveal how the CCD framework systematically balances the contribution of passive and active elements, leading to physically interpretable suspension designs that adapt to different driving behaviors.

\subsection{Performance Comparison}
Table~\ref{tab:returns_step3_susp} compares the system performance before and after the second CCD optimization. Two performance metrics are evaluated: (1) ride comfort, represented by the root mean square (RMS) of body vertical acceleration, and (2) control effort, computed as the mean of the absolute values of the four actuator forces. 

Although the RMS acceleration decreases only slightly for the mild driver (about 8\%), the control effort is reduced by more than 40\%. This outcome suggests that the second CCD optimization enhances control efficiency by refining the coordination between the suspension hardware and the control policy, achieving comparable comfort with substantially less actuation energy. 

For the aggressive driver, the RMS acceleration changes marginally, yet the control effort decreases by more than 50\%. This indicates that the optimized system effectively redistributes the control load among the actuators, exploiting the dynamics of the stiffer suspension to maintain performance while minimizing unnecessary high-frequency control actions. Overall, these results highlight the ability of the CCD framework to yield energy-efficient control without compromising ride quality.

\subsection{Analysis of Control Trajectories}
Figures~\ref{fig:action_traj_mild} and~\ref{fig:action_traj_aggr} illustrate the actuator force trajectories (\(u_1\)--\(u_4\)) before and after the second CCD optimization for mild and aggressive drivers, respectively. The dashed blue lines denote the actuator forces before optimization, and the solid orange lines represent those after optimization. Only the first 120 seconds of trajectories are shown for clarity. 

For the mild driver, the first actuator exhibits a pronounced reduction in force magnitude after the second CCD step, and the third and fourth actuators also show large decreases. These changes indicate smoother and more coordinated control actions, contributing to improved comfort and reduced energy consumption. 

For the aggressive driver, while the system still demands greater overall actuation to handle harsher maneuvers, most actuator trajectories become smoother with fewer abrupt peaks. The optimization thus preserves the responsiveness needed for aggressive driving while achieving better coordination among actuators. 

Overall, the second CCD step significantly reduces control magnitudes and smooths actuator responses for both driver types. These result demonstrate the framework’s capacity to adaptively tune physical and control designs to distinct driving behaviors.

\section{Conclusion}\label{sec6}
This work presents a multi-generation digital twin (DT)–based control co-design (CCD) framework for full-vehicle active suspension system to optimize the ride comfort and minimize the energy consumption. This framework integrates deep reinforcement learning (DRL), specifically the Proximal Policy Optimization (PPO) algorithm to jointly optimize the physical suspension components (spring stiffness and damping coefficient) and active controllers. By combining DTs and real-time updating the uncertainty-aware models, the framework enables the systems to learn and evolve with operational data, driver behaviors, and environmental conditions. With two distinct driving scenarios (mild and aggressive drivers), the case study demonstrates that the method can perform personalized optimization by tailoring both the physical parameters and control policies for different conditions and requirements. The multi-generation design process, where updated digital models informed subsequent CCD optimization, showcases how DTs can bridge virtual simulations and real-world systems to enhance adaptability, efficiency, and robustness across a vehicle’s lifecycle. These findings demonstrate how the DT framework enables active suspension systems that co-evolve physical system design with control strategies and self-improve through continual, data-driven learning.

Despite these promising results, the present study primarily focused on simulated environments with quantile learning of hypothetically real data. Future work will extend the framework to incorporate other advanced uncertainty quantification (UQ) and model update methods using data from real-world measurements, allowing the DT to capture stochastic variations and sensor noise in reality. In addition, the current implementation targeted suspension co-design only, while future research will generalize the CCD formulation to the entire autonomous vehicle by co-optimizing interconnected physical subsystems (e.g., transmission gear ratios, tire–road friction coefficients, and engine performance) with driving control strategies. Such an integrated, system-level co-design will advance the development of adaptive, high-performance, and safe autonomous vehicles that leverage DTs as learning-based decision-making systems throughout their operational lifecycles.

\backmatter

\bmhead{Supplementary information}

All codes used in this paper are freely available through  \href{https://github.com/TsaiYK/ControlCoDesign_DigitalTwin_RL_FullVehicleActiveSuspension}{the repository}.

\bmhead{Acknowledgements}

We are grateful for the grant support from the National Science Foundation (NSF)’s Engineering Research Center for Hybrid Autonomous Manufacturing: Moving from Evolution to Revolution (ERC-HAMMER) under Award Number EEC-2133630, NSF's FMRG: Manufacturing ADvanced Electronics through Printing Using Bio-based and Locally Identifiable Compounds (MADE-PUBLIC), under Award Number CMMI-2037026, and NASA's Center for In-space Manufacturing (CISM-R2): Recycling and Regolith Processing under Award Number 80NSSC24M0176. Yi-Ping Chen also acknowledges the Taiwan-Northwestern Doctoral Scholarship to support his doctoral study.

\section*{Declarations}
There are no conflicts of interest.

\printbibliography

\end{document}